\documentclass[10pt,twocolumn,letterpaper]{article}
\usepackage[pagenumbers]{cvpr}
\usepackage{amsmath,amssymb,amsthm}
\usepackage{array,multirow}
\usepackage{microtype}
\usepackage{tikz}
\usetikzlibrary{arrows.meta,positioning,calc,fit}
\usepackage{enumitem}
\usepackage{placeins}
\usepackage{stfloats}
\usepackage{booktabs}
\definecolor{modelblue}{RGB}{35,83,135}
\definecolor{motionorange}{RGB}{196,112,27}
\definecolor{effectgreen}{RGB}{29,117,100}
\definecolor{draftred}{RGB}{156,39,48}

\newcommand{\method}{MotionPort}

\definecolor{cvprblue}{rgb}{0.21,0.49,0.74}
\usepackage[pagebackref,breaklinks,colorlinks,allcolors=cvprblue]{hyperref}
\usepackage[capitalize,nameinlink,noabbrev]{cleveref}
\def\confName{CVPR}
\def\confYear{2027}

\title{What Should World Models Plan Over? \\
Learning Geometry-Aligned Representations of Body Motion}

\author{Renhao Zhang\\
University of Massachusetts\\
Amherst, MA, USA\\
{\tt\small renhaozhang@cs.umass.edu}
\and
Abhranil Chandra\\
University of Massachusetts\\
Amherst, MA, USA\\
{\tt\small abhranilchan@cs.umass.edu}
\and
Bruno Castro da Silva\\
University of Massachusetts\\
Amherst, MA, USA\\
{\tt\small bsilva@cs.umass.edu}
}
\hypersetup{pdftitle={What Should World Models Plan Over? Learning Geometry-Aligned Representations of Body Motion},pdfauthor={Renhao Zhang, Abhranil Chandra, Bruno Castro da Silva}}

\begin{document}
\maketitle
\newcommand{\teasermaxheight}{.28\textheight}
\newcommand{\MotionPortTeaserCaption}{%
\textbf{Body motion connects prediction and planning.}
(a) MotionPort explicitly represents body motion as an intermediate step between actions and their effects on the environment, rather than encoding action chunks.
(b) Illustrative optimization in latent macro-action space versus geometry-structured motion-code space.
(c) MotionPort maintains high success on short- and long-term PushT goals.
}

\providecommand{\teasermaxheight}{.32\textheight}
\begin{figure*}[!b]
\centering
\includegraphics[width=\textwidth,height=\teasermaxheight,keepaspectratio]{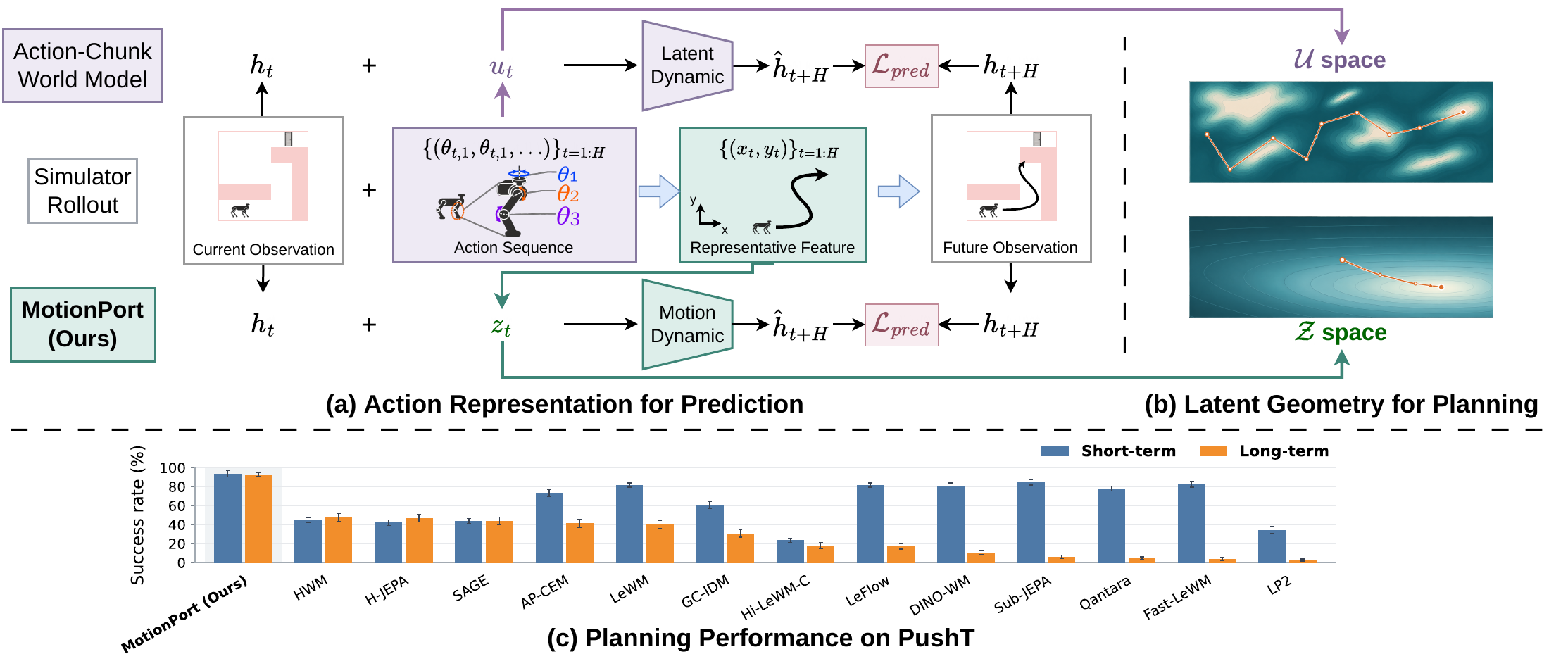}
\caption{\MotionPortTeaserCaption}
\label{fig:teaser}
\end{figure*}

\begin{abstract}
We introduce MotionPort, which learns low-dimensional latent representations of body motion for accurate dynamics prediction and effective long-horizon planning. Existing world-model planners often search over action representations learned for prediction, yet we show that predictive accuracy alone does not ensure that optimization can discover successful trajectories. MotionPort addresses this mismatch by representing body trajectories rather than action chunks and organizing the search space by motion similarity, helping predicted improvements translate into successful execution. Without fine-tuning state encoders, MotionPort raises average goal-reaching success from 37.9\% to 87.8\% across three backbones and seven settings in four domains. On the evaluated AntMaze–OGBench tasks, it achieves 72\% success, compared with 4\% for the strongest baseline.
\end{abstract}

\section{Introduction}
\label{sec:introduction}

Visual world models enable goal-directed planning by predicting how candidate actions affect the environment~\cite{dinowm,lewm,pldm}. Extending this process to longer horizons, however, introduces two challenges: prediction errors accumulate across imagined steps, and searching over extended sequences of primitive actions becomes increasingly costly~\cite{hwm,hjepa}. Recent works address these challenges through temporal abstraction, compressing primitive action sequences into latent macro-actions and predicting at coarser timescales~\cite{hwm,hjepa,oposm}. As illustrated in Fig.~\ref{fig:teaser}(c), some approaches employing temporal abstraction exhibit relatively small performance drops from short- to long-horizon goals. 
Beyond temporal compression, these macro-action codes condition the predictive dynamics and define the planner’s search space.

These two roles, however, impose different requirements on the same latent space. Macro-action codes are typically learned to predict future states from recorded trajectories, whereas gradient-based planners optimize these codes to minimize goal-dependent costs~\cite{train_test_gap,hjepa}. Yet accurate prediction does not guarantee an optimization-friendly search space~\cite{jepa_fidelity,drpe}: the resulting cost landscape may contain flat regions or poor local minima that hinder gradient-based search~\cite{train_test_gap,dinowm}. Moreover, even when the predicted cost decreases, the planner may simply exploit errors in the learned dynamics rather than discover actions that lead to genuine progress in the environment~\cite{model_exploitation,train_test_gap}. This raises a fundamental question: \textit{can we learn an action representation that supports accurate prediction while providing a more effective search space for goal-directed planning?}

We introduce \textbf{MotionPort}, which organizes world-model prediction and planning around the agent's own body motion. As illustrated in Fig.~\ref{fig:teaser}(a), in many physics-based simulators, low-level actions induce movements of the agent's body, which in turn interact with the surrounding environment to determine future states~\cite{mujoco}. For example, an Ant’s joint actions drive coordinated body motion, while the simulator resolves interactions with the maze, including collisions with walls, to determine the resulting state~\cite{hjepa,mujoco}. Conventional action-chunk world models instead predict future states directly from encoded action sequences, leaving this intermediate body motion implicit~\cite{dinowm,hwm,hjepa}. MotionPort makes it explicit by encoding short trajectories of representative body features into compact motion codes, without directly encoding the underlying actions. We further train a conditional generative policy to translate these codes into executable actions given the current body state~\cite{pdp,flowmatching}, and a motion-conditioned dynamics model~\cite{skimo,oposm} to predict their consequences in the environment. Beyond prediction, body motion also provides a natural structure for planning. For goal-reaching tasks, similar body trajectories from the same initial state tend to produce nearby outcomes and comparable goal costs. We therefore align distances between motion codes with differences in their corresponding body trajectories~\cite{pdp}, encouraging local updates to produce coherent changes in physical behavior, as illustrated in Fig.~\ref{fig:teaser}(b). To make planning in this space more effective, we additionally learn motion priors to provide plausible initial candidates~\cite{sage,worldplanner} and a temporal goal metric to evaluate their predicted progress toward distant goals~\cite{hilp,mbold,tempo,time_to_goal}. The planner then optimizes these motion codes through imagined rollouts and decodes the selected motions into actions for execution. Our contributions are:
\begin{itemize}[leftmargin=*,itemsep=2pt,topsep=3pt]
    \item We introduce MotionPort, which represents body trajectories rather than action chunks in a low-dimensional latent space for world-model prediction and planning. Aligning distances between motion codes with distances between the corresponding body trajectories encourages planning updates to produce coherent changes in physical behavior.

    \item We show that predictive accuracy alone does not ensure effective planning, and that motion-space geometry makes learned priors more effective for discovering successful trajectories. Together with temporal goal scoring, this structure helps translate improvements in imagined rollouts into physical progress.

    \item We demonstrate that MotionPort improves planning without fine-tuning pretrained state encoders. Across LeWM, HWM, and H-JEPA~\cite{lewm,hwm,hjepa}, it improves goal-reaching success in all 21 comparisons spanning seven settings in four domains, with an average gain of 50 percentage points. On the evaluated AntMaze–OGBench tasks, it achieves 72\% success, compared with 4\% for the strongest baseline.
\end{itemize}

\section{Related Work}

\paragraph{Visual world models and action abstraction.} Visual world models enable planning by predicting action-conditioned futures in learned representations. DINO-WM~\cite{dinowm} predicts pretrained visual features, while LeWM~\cite{lewm} and PLDM~\cite{pldm} learn predictive representations from offline trajectories. To extend planning over longer horizons, skill-based approaches such as SkiMo~\cite{skimo} and OPOSM~\cite{oposm} learn temporally abstract dynamics, while HWM~\cite{hwm}, H-JEPA~\cite{hjepa}, and Hi-LeWM~\cite{hilewm} introduce hierarchical world models with latent macro-actions. Related work also explores the structure of behavior representations: PDP~\cite{pdp} learns geometry-aligned behavior latents for controllable action generation and test-time adaptation.

\paragraph{Planning reliability and goal-directed search.} Accurate prediction on recorded trajectories does not necessarily translate into effective planning, as studied through decision-relevant prediction errors~\cite{drpe}, JEPA prediction fidelity~\cite{jepa_fidelity}, and model exploitation~\cite{model_exploitation}. Closing the Train-Test Gap~\cite{train_test_gap} addresses this mismatch through planning-aware world-model training, while Temporal Straightening~\cite{temporal_straightening} improves gradient-based planning by regularizing state-representation geometry. Other approaches improve planning through structured search and candidate generation, including WorldPlanner~\cite{worldplanner}, SAGE~\cite{sage}, and LeFlow~\cite{leflow}, or through learned temporal and functional goal distances, including HILP~\cite{hilp}, MBOLD~\cite{mbold}, TEMPO~\cite{tempo}, and Time-to-Goal~\cite{time_to_goal}. These works primarily address prediction reliability, state representations, candidate selection, or goal evaluation, leaving the organization of executable action representations as a complementary direction.

\begin{figure*}[t]
\centering
\includegraphics[width=\textwidth,height=.24\textheight,keepaspectratio]{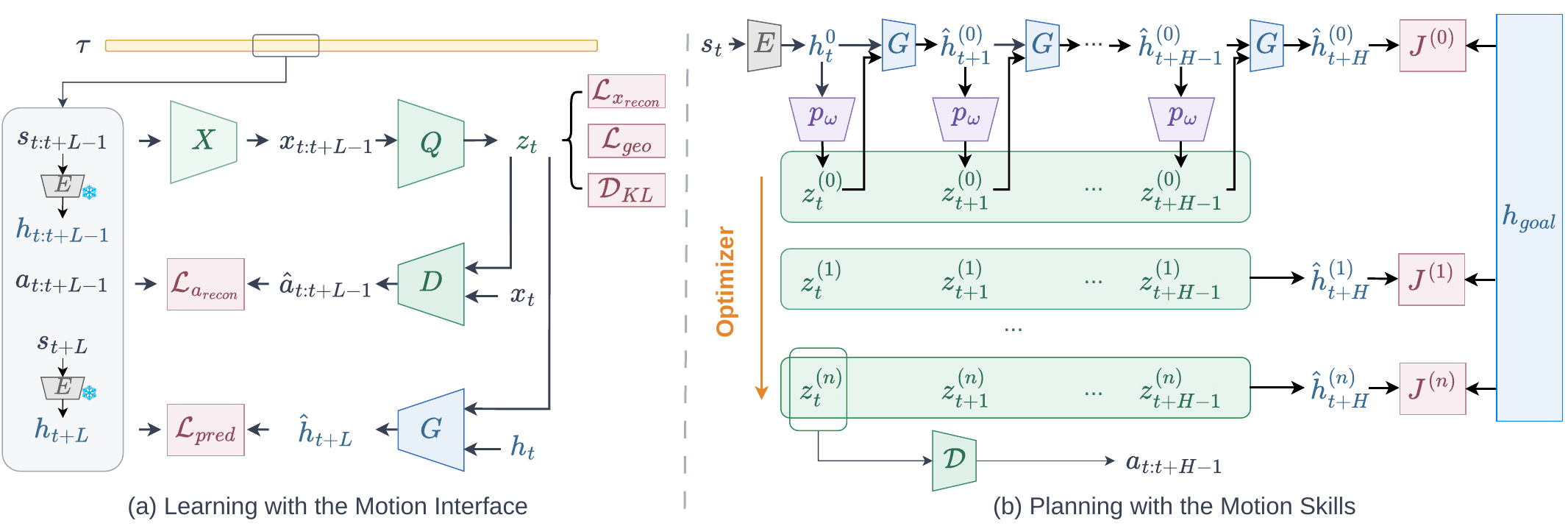}
\caption{\textbf{Learning and planning with MotionPort.} (a) Representative body trajectories are encoded into geometry-structured motion codes, which condition an action decoder and a dynamics model, while the pretrained state encoder remains frozen. (b) A learned prior initializes motion-code sequences, which are iteratively refined by an optimizer. The diagram shows successive motion segments horizontally and iterative plan refinement vertically. The selected motion is decoded into executable actions and checked before execution.}
\label{fig:method_overview}
\end{figure*}

\section{MotionPort}
\label{sec:method}

MotionPort focuses on the representation of behavior for world-model prediction and planning, while preserving the visual state representations learned by existing world models~\cite{lewm,hwm,hjepa}. Specifically, we use a pretrained state encoder $E$, which remains frozen and maps the current observation $s_t$ and goal observation $s_g$ to latent features $h_t=E(s_t)$ and $h_g=E(s_g)$. Each observation contains an image $I_t$ and available body state $o_t$, such as joint states or end-effector poses. Given an offline dataset $\mathcal{D}$ of state--action trajectories, we learn compact motion codes $z$ that represent body movements spanning $L$ primitive actions. We then train a motion-conditioned dynamics model to predict how these movements affect the environment, and a conditional generative policy to translate motion codes into executable actions. At test time, the planner optimizes a sequence $Z=(z_1,\ldots,z_H)$ of $H$ motion codes while keeping all model parameters fixed. Figure~\ref{fig:method_overview} provides an overview of the learning and planning procedures.

\subsection{Geometry-Structured Motion Codes}
\label{sec:method_skills}

We represent the agent's movement using representative body features rather than the actions that produce it~\cite{pdp}. Specifically, a fixed function $X$ extracts motion-related features $x_t=X(s_t)$, such as end-effector poses, pusher positions, or locomotion coordinates. For a segment of $L$ primitive actions, we obtain a body-motion trajectory $x=x_{t:t+L-1}$, expressed relative to its starting configuration when appropriate. As shown in Fig.~\ref{fig:method_overview}(a), we learn a variational motion encoder $Q_\psi$ from these trajectories, without directly encoding primitive actions, scene images, or task goals. The encoder defines $q_\psi(z\mid x)=\mathcal{N}(\mu_\psi(x),\Sigma_\psi(x))$, with diagonal covariance $\Sigma_\psi(x)$, while a reconstruction head $R_\rho$ recovers the normalized motion features $\tilde{x}$. The training objective combines three weighted losses:

\begin{equation}
\begin{aligned}
\mathcal{L}_{\mathrm{motion}} ={}&
\lambda_{\mathrm{rec}}\mathcal{L}_{\mathrm{x_{recon}}}
+\lambda_{\mathrm{geo}}\mathcal{L}_{\mathrm{geo}}+\lambda_{\mathrm{KL}}\mathcal{L}_{\mathrm{KL}}.
\end{aligned}
\label{eq:motion-loss}
\end{equation}

Here, $\mathcal{L}_{\mathrm{x_{recon}}}=\mathbb{E}_{x,\,z\sim q_\psi(\cdot\mid x)}[\mathrm{MSE}(R_\rho(z),\tilde{x})]$ preserves motion information, while $\mathcal{L}_{\mathrm{KL}}=\mathbb{E}_x[D_{\mathrm{KL}}(q_\psi(\cdot\mid x)\Vert\mathcal{N}(0,I))]$ regularizes the latent distribution. However, reconstruction alone does not explicitly organize the latent space according to physical similarity. Since gradient-based planning continuously modifies motion codes, we align distances between codes with differences in their corresponding body trajectories. Given a minibatch $\mathcal{B}=\{x_i\}_{i=1}^{B}$, let $\mathcal{I}$ denote a set of sampled index pairs $(i,j)$ with $1\leq i<j\leq B$. We define the geometry loss as

\begin{equation}
\mathcal{L}_{\mathrm{geo}}=\frac{1}{|\mathcal{I}|}\sum_{(i,j)\in\mathcal{I}}\left(\frac{\delta_z(i,j)}{r_z}-\frac{\delta_x(i,j)}{r_x}\right)^2.
\label{eq:geometry-loss}
\end{equation}

Here, $\delta_z(i,j)=\|\mu_\psi(x_i)-\mu_\psi(x_j)\|_2^p$ and $\delta_x(i,j)$ measures the corresponding distance between scaled body trajectories using the same norm power $p$. The factors $r_z$ and $r_x$ normalize the respective distances over the sampled pairs. This constraint encourages local changes in motion codes to correspond to coherent changes in physical behavior without assuming identical environmental effects. After training, we use the encoder mean, standardized using training-set statistics, as the deterministic motion code $z$ for subsequent prediction and action generation. Further analyses of motion-code geometry are provided in Appendices~A and~C.2.

\subsection{Predicting and Executing Motions}
\label{sec:method_dynamics}

With the motion encoder $Q_\psi$ fixed, we learn two complementary mappings from motion codes: generating the actions required to realize a movement and predicting its consequences in the environment. To generate executable actions, we train a conditional generative policy $D_\theta$ using flow matching~\cite{flowmatching}. Given the current body state $o_t$ and motion code $z_t$, the policy generates an action chunk $\hat{a}_{t:t+L-1}=D_\theta(o_t,z_t)$. For a normalized training action chunk $\tilde{a}$, Gaussian noise $\epsilon$, and flow time $\xi\in[0,1]$, we define $y_\xi=(1-\xi)\epsilon+\xi\tilde{a}$ and optimize
\begin{equation}
\mathcal{L}_{a_{\mathrm{recon}}}=\mathbb{E}_{\mathcal{B},\epsilon,\xi}\left[\mathrm{MSE}\left(v_\theta(y_\xi,\xi,o_t,z_t),\tilde{a}-\epsilon\right)\right].
\label{eq:action-recon}
\end{equation}
Here, $v_\theta$ is the conditional flow velocity network, whose integration produces the decoded action chunk. 
We train a motion-conditioned dynamics model $G_\eta$ to predict future visual features from motion codes, using the frozen $E$.
Rather than predicting only the endpoint of a motion segment, $G_\eta$ also predicts future visual representations at intermediate steps. Let $\mathcal{K}\subseteq\{1,\ldots,L\}$ denote the prediction offsets within a segment, including its endpoint $L$. For each $k\in\mathcal{K}$, the model predicts $\hat{h}_{t+k}=G_{\eta,k}(h_t,o_t,z_t)$, supervised by the corresponding future representation through
\begin{equation}
\mathcal{L}_{\mathrm{pred}}=\mathbb{E}_{\mathcal{B},k\in\mathcal{K}}\left[\mathrm{MSE}\left(\hat{h}_{t+k},h_{t+k}\right)\right].
\label{eq:motion-pred}
\end{equation}
During planning, these predictions are composed across motion segments to support long-horizon rollouts, while intermediate predictions accommodate shorter execution prefixes. Additional details are provided in Appendix~D.1.

\subsection{Planning in Motion Space}
\label{sec:motion-planning}

To initialize planning, we learn goal-conditioned and goal-free motion priors $p_\omega^g(z\mid h,o,h_g)$ and $p_\omega^0(z\mid h,o)$ using conditional flow matching~\cite{flowmatching} on recorded motion codes, with future observations providing goal labels. Rather than generating entire sequences directly, we construct each candidate by alternating motion-code sampling and dynamics prediction, as illustrated in Fig.~\ref{fig:method_overview}(b). Starting from $(\hat{h}_0,\hat{o}_0)=(h_t,o_t)$, each motion code $z_j$ is sampled from a mixture of the goal-conditioned prior, goal-free prior, and Gaussian distribution. The dynamics model then predicts the next state $\hat{h}_j=G_{\eta,L}(\hat{h}_{j-1},\hat{o}_{j-1},z_j)$, with the body state $\hat{o}_j$ updated accordingly. Repeating this process for $H$ segments produces $N$ initial candidate sequences $\{Z_i^{(0)}\}_{i=1}^{N}$.

To evaluate progress toward a goal beyond visual proximity, we learn two state embeddings $\phi_\ell$, $\ell\in\{1,2\}$, following HILP-inspired expectile temporal-difference learning~\cite{hilp}. Each embedding maps a single visual feature $h$ to a latent vector, while the distance between two embeddings defines a goal-conditioned value $V_\ell(h,h_g)=-\|\phi_\ell(h)-\phi_\ell(h_g)\|_2$. For a recorded transition $h\rightarrow h'$, we relabel $h_g$ as the current state, a future state in the same trajectory, or a randomly sampled training state. Let $m=0$ when the current and goal states correspond to the same dataset node, and $m=1$ otherwise. Using target value networks $\bar{V}_\ell$, we construct TD targets $y_\ell=-m+\gamma m\bar{V}_\ell(h',h_g)$ and minimize
\begin{equation}
\mathcal{L}_{\phi}=\mathbb{E}\left[w_\tau(a_{\mathrm{TD}})\sum_{\ell=1}^{2}\left(V_\ell(h,h_g)-\mathrm{sg}[y_\ell]\right)^2\right],
\label{eq:temporal-td}
\end{equation}
where $a_{\mathrm{TD}}=\min_\ell y_\ell-\frac{1}{2}\sum_{\ell=1}^{2}\bar{V}_\ell(h,h_g)$, $\mathrm{sg}$ denotes stop-gradient, and $w_\tau(a)$ assigns weight $\tau$ to nonnegative advantages and $1-\tau$ otherwise. This objective propagates goal-reaching information across transitions, allowing the learned distances to reflect temporal reachability rather than visual proximity alone. At planning time, we define $d_\phi(h,h_g)=\frac{1}{2}\sum_{\ell=1}^{2}\|\phi_\ell(h)-\phi_\ell(h_g)\|_2$ and combine it with visual distance:
\begin{equation}
c(h,h_g)=(1-w_\phi)\frac{\|h-h_g\|_2^2}{r_h}+w_\phi\frac{d_\phi(h,h_g)}{r_\phi},
\label{eq:goal-cost}
\end{equation}
where $w_\phi$ balances the two distances and $r_h,r_\phi$ are fixed normalization factors.
We refine each candidate motion sequence $Z$ by minimizing
\begin{equation}
J(Z)=\sum_{j=1}^{H}\alpha_j c(\hat{h}_j,h_g)+\lambda_{\mathrm{dis}}\Delta_G(Z),
\label{eq:planning-objective}
\end{equation}
where $\alpha_H=1$ assigns full weight to the predicted endpoint, while intermediate weights $\alpha_j$ optionally encourage progress along the trajectory. When the remaining execution budget ends within a motion segment, we use its latest valid prefix prediction as the endpoint. To account for prediction uncertainty, we independently train a second motion-conditioned dynamics model using the same prediction objective as $G_\eta$. Both models predict the outcomes of a candidate motion sequence, and $\Delta_G(Z)$ measures their disagreement in predicted visual features, normalized by a fixed calibration scale. Larger disagreement indicates less consistent predictions and is penalized during planning. We jointly optimize all motion codes in $Z$ using AdamW, keeping both dynamics models fixed throughout optimization.

After refining the candidate sequences, we perform a final execution check using the pretrained, frozen action-conditioned predictor $F_1$, without further optimization. Specifically, we select the $m$ lowest-cost sequences and decode their first motion codes into executable action chunks. The best sequence under $J$ defines a common subgoal $h_{\mathrm{sub}}$, corresponding to its first-motion endpoint predicted by $G_\eta$. For each shortlisted candidate $i\in\mathcal{C}$, we use $F_1$ to predict the outcome of its decoded actions, obtaining $\tilde{h}_i$ in the planning feature space. We then select the candidate whose predicted outcome is closest to the common subgoal:
\begin{equation}
i_{\mathrm{exec}}=\arg\min_{i\in\mathcal{C}}\mathrm{MSE}(\tilde{h}_i,h_{\mathrm{sub}}).
\label{eq:execution-selection}
\end{equation}
This step only re-ranks the shortlisted candidates without updating their motion codes or actions. We execute a prefix of the selected action chunk and replan from the next observation. Further details are provided in Appendix~D.

\section{Experiments}
\label{sec:experiments}
\raggedbottom
\widowpenalty=10000
\clubpenalty=10000
\makeatletter
\setlength{\@fptop}{0pt}
\setlength{\@dblfptop}{0pt}
\setlength{\@dblfpsep}{12pt}
\makeatother

In this section, we evaluate MotionPort's goal-reaching performance and planning efficiency, and examine the roles of motion geometry, learned proposals, and temporal goal scoring. We consider seven evaluation settings across four environments, following the visual-control protocols of H-JEPA~\cite{hjepa} where applicable. PushT and Cube retain their standard task protocols, with additional goals sampled under the same settings. FourRoom includes a standard setting requiring two doorway crossings and a hard setting requiring three, with longer detours under the same execution budget. Visual AntMaze includes a standard setting with the original D3 goals, a hard setting mixing new D3 and longer D6 goals under the same budget, and an OGBench setting with official long-range goals and a stricter success threshold. MotionPort can be integrated with different pretrained state encoders without fine-tuning them. We focus our detailed analyses on HWM, with cross-backbone comparisons in \cref{sec:exp_backbones} and complete results in Appendix~C.5.

\begin{table*}[t]
\caption{\textbf{Goal-reaching success on Push-T, Cube, FourRoom, and AntMaze.} Results show mean success rates (\%) and standard errors across seven settings. MotionPort uses a frozen HWM encoder. $\Delta$ denotes improvement over the strongest baseline in each setting.}
\label{tab:main_control}
\centering
\begingroup
\small
\setlength{\tabcolsep}{2pt}
\renewcommand{\arraystretch}{1.12}
\newcommand{\mainstat}[2]{#1\,{\normalfont\scriptsize$(\pm #2)$}}
\begin{tabular*}{\textwidth}{@{\extracolsep{\fill}}lcccccccc@{}}
\toprule
\multirow{2}{*}{Method} & \multirow{2}{*}{PushT} & \multirow{2}{*}{Cube} & \multicolumn{2}{c}{FourRoom} & \multicolumn{3}{c}{AntMaze} & \multirow{2}{*}{Average} \\
\cmidrule(lr){4-5}\cmidrule(lr){6-8}
 & & & standard & hard & standard & hard & OGBench & \\
\midrule
H-JEPA & \mainstat{46.67}{4.10} & \mainstat{49.33}{4.08} & \mainstat{98.00}{2.00} & \mainstat{41.00}{4.94} & \mainstat{78.00}{5.92} & \mainstat{22.82}{4.94} & \mainstat{0.00}{0.00} & 47.97 \\
HWM & \mainstat{47.33}{4.07} & \mainstat{61.33}{4.00} & \mainstat{98.00}{2.00} & \mainstat{23.00}{4.23} & \mainstat{60.00}{7.00} & \mainstat{14.00}{4.29} & \mainstat{0.00}{0.00} & 43.38 \\
LeWM & \mainstat{40.00}{4.03} & \mainstat{30.67}{3.79} & \mainstat{58.00}{7.05} & \mainstat{0.00}{0.00} & \mainstat{24.00}{6.10} & \mainstat{3.00}{1.71} & \mainstat{0.00}{0.00} & 22.24 \\
SAGE & \mainstat{44.00}{4.07} & \mainstat{68.67}{3.79} & \mainstat{98.00}{2.00} & \mainstat{80.00}{4.02} & \mainstat{20.00}{5.71} & \mainstat{11.00}{3.14} & \mainstat{4.00}{4.00} & 46.52 \\
LeFlow & \mainstat{17.33}{3.11} & \mainstat{92.00}{0.67} & \mainstat{0.00}{0.00} & \mainstat{0.00}{0.00} & \mainstat{2.00}{2.00} & \mainstat{0.00}{0.00} & \mainstat{0.00}{0.00} & 15.90 \\
Hi-LeWM-C & \mainstat{18.00}{3.15} & \mainstat{4.00}{1.61} & \mainstat{38.00}{6.93} & \mainstat{0.00}{0.00} & \mainstat{18.00}{5.49} & \mainstat{0.00}{0.00} & \mainstat{0.00}{0.00} & 11.14 \\
\textbf{MotionPort} & \mainstat{\textbf{92.67}}{2.13} & \mainstat{\textbf{97.33}}{1.31} & \mainstat{\textbf{100.00}}{0.00} & \mainstat{\textbf{100.00}}{0.00} & \mainstat{\textbf{92.00}}{3.88} & \mainstat{\textbf{58.00}}{4.96} & \mainstat{\textbf{72.00}}{10.20} & \textbf{87.43} \\
\midrule
$\Delta$ vs. best & +45.33 & +5.33 & +2.00 & +20.00 & +14.00 & +35.18 & +68.00 & +39.45 \\
\bottomrule
\end{tabular*}
\endgroup
\end{table*}

\paragraph{Evaluation protocol.} We evaluate goal-reaching performance using \emph{ever success}, which measures whether the agent reaches the goal at any point during an episode. We also report \emph{terminal success} at the final execution step in \cref{sec:exp_retention}. For each evaluation setting, we report the mean success rate across all evaluation tasks, with error bars indicating standard errors. Additional details on task configurations and evaluation protocols are provided in Appendices~A, C.1, and~C.5.

\subsection{MotionPort Performance Across Seven Settings}
\label{sec:exp_control}

We compare MotionPort against six world-model planning baselines: LeWM, HWM, H-JEPA, SAGE, LeFlow, and Hi-LeWM-C~\cite{lewm,hwm,hjepa,sage,leflow,hilewm}. All methods are evaluated under the same task settings and environment-step budgets, with MotionPort using a frozen HWM state encoder.

As shown in \cref{tab:main_control}, MotionPort achieves the highest mean success rate across all seven settings. It reaches 92.67\% on PushT, compared with 47.33\% for the strongest baseline, and achieves 100\% on hard FourRoom, outperforming the previous best result of 80\%. The improvement is particularly pronounced on AntMaze--OGBench, where MotionPort achieves 72\% success compared with 4\% for the strongest baseline. These results demonstrate effective goal-reaching across both manipulation and navigation, including challenging long-range tasks.

\FloatBarrier
\subsection{From Prediction to Physical Execution}
\label{sec:exp_reliability}\label{sec:exp_limits}\label{sec:exp_prediction}

High goal-reaching success does not necessarily imply that a planner's predictions remain reliable under optimization. A world model is trained on logged trajectories, but planning evaluates commands selected to minimize its predicted cost. We investigate this discrepancy on PushT and Cube by comparing prediction accuracy on logged trajectories and planner-selected commands, followed by the physical outcomes of latent-space optimization.
We compare the native HWM macro-action planner (\emph{HWM-Macro}), MotionPort without geometry (\emph{MP-noGeo}), and full MotionPort using frozen HWM encoders. MP-noGeo differs from MotionPort only by setting $\lambda_{\mathrm{geo}}=0$ in \cref{eq:motion-loss}, with all other training and planning settings unchanged.

We first compare prediction accuracy on logged trajectories and planner-selected actions. As shown in \cref{tab:planning_reliability}, HWM-Macro and MotionPort have similar validation errors on PushT (2.81 vs. 2.89), while HWM-Macro achieves lower validation error on Cube (0.42 vs. 0.68\,cm). However, its planning errors are substantially higher, reaching 19.00 on PushT and 5.69\,cm on Cube, compared with 3.43 and 1.07\,cm for MotionPort. Accordingly, the error ratios (ER) are 6.76 and 13.65 for HWM-Macro, but only 1.19 and 1.59 for MotionPort. MP-noGeo also achieves low planning errors, but falls behind full MotionPort in success rate.
These results show that accurate prediction on logged trajectories does not necessarily translate into reliable predictions on planner-selected actions (Appendix~C.1).

\begin{table}[t]
\caption{\textbf{Prediction reliability and control performance on PushT and Cube.} Validation error (VE) measures prediction accuracy on logged trajectories, while planning error (PE) measures prediction accuracy on planner-selected actions. Error ratio (ER) is defined as PE/VE, and success rate (SR) measures goal-reaching performance (\%). Results are reported as mean $\pm$ standard error using frozen HWM42. Prediction errors are measured in latent SSE for PushT and position error (cm) for Cube.}
\label{tab:planning_reliability}
\centering\small
\begingroup
\setlength{\tabcolsep}{1.5pt}
\renewcommand{\arraystretch}{1.10}
\newcommand{\predstat}[2]{#1\,{\normalfont\scriptsize$(\pm #2)$}}
\begin{tabular*}{\columnwidth}{@{\extracolsep{\fill}}llrrr@{}}
\toprule
Domain & Metric & HWM-Macro & MP-noGeo & MotionPort \\
\midrule
\multirow{4}{*}{PushT} & VE & \predstat{2.81}{0.42} & \predstat{2.80}{0.48} & \predstat{2.89}{0.49} \\
 & PE & \predstat{19.00}{1.45} & \predstat{3.39}{0.23} & \predstat{3.43}{0.25} \\
 & ER & \predstat{6.76}{1.13} & \predstat{1.21}{0.22} & \predstat{1.19}{0.22} \\
 & SR & \predstat{47.33}{4.07} & \predstat{71.33}{2.31} & \predstat{92.67}{2.13} \\
\midrule
\multirow{4}{*}{Cube} & VE & \predstat{0.42}{0.01} & \predstat{0.70}{0.01} & \predstat{0.68}{0.01} \\
 & PE & \predstat{5.69}{0.19} & \predstat{1.38}{0.06} & \predstat{1.07}{0.04} \\
 & ER & \predstat{13.65}{0.53} & \predstat{1.97}{0.09} & \predstat{1.59}{0.06} \\
 & SR & \predstat{61.33}{4.00} & \predstat{85.33}{1.73} & \predstat{97.33}{1.31} \\
\bottomrule
\end{tabular*}
\endgroup
\end{table}

\begin{figure*}[t]
\centering
\includegraphics[width=\textwidth,height=.64\textheight,keepaspectratio]{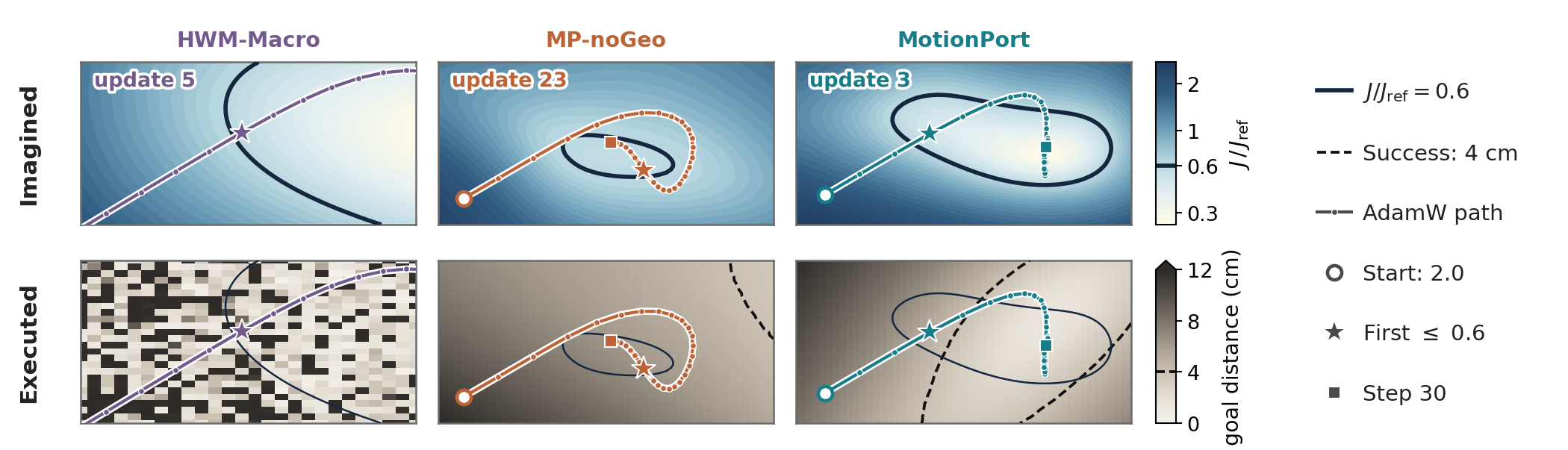}
\caption{\textbf{From imagined optimization to physical execution on Cube.} We visualize two-dimensional slices of the planning spaces of HWM-Macro, MP-noGeo, and MotionPort. The top row shows predicted goal costs, while the bottom row shows actual cube-to-goal distances after executing the corresponding actions for 20 environment steps. The same optimization path is shown in both rows, with circles marking the starting points, stars marking the first updates reaching the predicted-cost threshold, and squares marking the final solutions after 30 updates. Solid contours indicate the predicted low-cost region ($J/J_{\mathrm{ref}}\leq 0.6$), while dashed contours indicate the physical success region (distance $\leq 4$\,cm). MotionPort exhibits a larger overlap between these regions, allowing optimization to find a solution that succeeds.}
\label{fig:latent_optimization}
\end{figure*}

The large gap between validation and planning errors raises a question: is HWM's dynamics model inherently unreliable, or does its planner select actions for which predictions are inaccurate? Starting from the same 100 Cube states, we run HWM-Macro and MotionPort and obtain executable action sequences through HWM's low-level solver and MotionPort's motion decoder, respectively. We feed both sets of primitive actions into HWM's original action-conditioned predictor $F_1$ and compare its predictions with the visual features observed after ten environment steps. The sum of squared errors is 52.71 for HWM-Macro's actions but only 1.59 for MotionPort's (Appendix~C.1). Thus, the same predictor is substantially more accurate on MotionPort's actions than on those selected by HWM's own planner. This suggests that optimization can steer the planner toward actions for which the model's predictions are unreliable. Such actions may appear promising because of prediction errors, allowing the optimizer to lower the imagined cost without making corresponding physical progress---a form of \emph{model exploitation}. But why does the planning space allow such misleading improvements?

\begin{figure*}[t]
\centering
\includegraphics[width=\textwidth,clip]{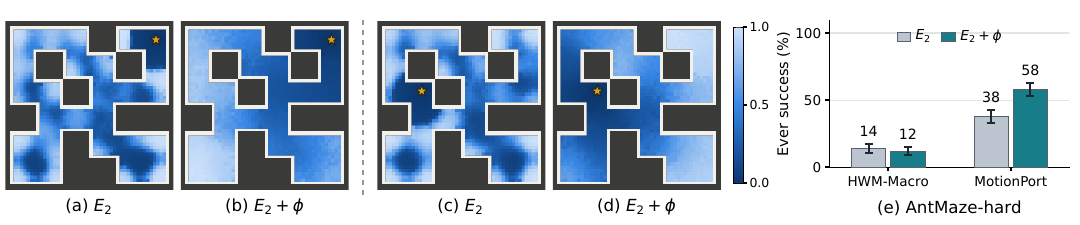}
\caption{\textbf{A better goal score benefits motion-space planning.} (a--d) Goal-cost landscapes for two AntMaze goals, comparing visual-feature distance ($E_2$) with the combined visual and temporal score ($E_2+\phi$). Darker regions indicate lower costs, and stars mark the goals. The learned temporal score better captures the maze's reachable paths. (e) Goal-reaching success on AntMaze-hard with and without $\phi$, where \method{} improves from 38\% to 58\%, while HWM-Macro decreases from 14\% to 12\%.}
\label{fig:phi_score_landscape}
\end{figure*}

Figure~\ref{fig:latent_optimization} illustrates this problem by comparing imagined costs and actual execution outcomes in two-dimensional slices of the planning spaces on Cube. The top row shows the predicted goal cost, while the bottom row shows the actual cube-to-goal distance after executing the corresponding actions. The same optimization path is shown in both rows, allowing us to see whether reducing the predicted cost also improves the real outcome. HWM-Macro reaches the low-cost region after only five updates, but nearby points in its planning space lead to drastically different physical outcomes. This explains how the optimizer can exploit prediction errors: actions that appear promising to the model do not necessarily bring the cube closer to its goal. In contrast, MP-noGeo and MotionPort directly optimize representations of body motion (\cref{sec:method_skills}) rather than abstract action chunks, producing much smoother execution maps. However, their optimization landscapes differ substantially. For MP-noGeo, the predicted low-cost region is relatively small and poorly aligned with the physical success region, leaving the final solution outside the goal. With geometry regularization (\cref{eq:geometry-loss}), MotionPort exhibits a broader low-cost region that overlaps much more with the actual success region. So, its optimizer has a larger region of physically successful solutions to reach, and the final solution achieves the goal in this example. These results illustrate how structuring the planning space around physical motion helps turn predicted improvements into successful execution, rather than simply minimizing an imperfect model's cost.

\setcounter{topnumber}{1}
\begin{table}[t]
\caption{\textbf{Effect of learned priors on goal-reaching.} NP (No Prior) uses Gaussian initialization, while LP (Learned Prior) uses a 1:1 mixture of prior and Gaussian samples. $\Delta$P denotes the improvement from NP to LP in percentage points.}
\label{tab:prior_closed_loop}\label{tab:planning_components}
\centering\small
\begingroup
\setlength{\tabcolsep}{2pt}
\renewcommand{\arraystretch}{1.10}
\newcommand{\priorstat}[2]{#1\,{\normalfont\scriptsize$(\pm #2)$}}
\begin{tabular*}{\columnwidth}{@{\extracolsep{\fill}}llrrr@{}}
\toprule
Domain & Metric & HWM-Macro & MP-noGeo & MotionPort \\
\midrule
\multirow{3}{*}{PushT} & NP & \priorstat{47.3}{4.1} & \priorstat{62.6}{2.3} & \priorstat{80.9}{2.5} \\
 & LP & \priorstat{55.8}{2.8} & \priorstat{71.3}{2.3} & \priorstat{92.7}{2.1} \\
 & $\Delta$P & +8.5 & +8.7 & +11.8 \\
\midrule
\multirow{3}{*}{Cube} & NP & \priorstat{61.3}{4.0} & \priorstat{57.1}{2.3} & \priorstat{58.4}{2.7} \\
 & LP & \priorstat{71.1}{2.5} & \priorstat{85.3}{1.7} & \priorstat{97.3}{1.3} \\
 & $\Delta$P & +9.8 & +28.2 & +38.9 \\
\midrule
\multirow{3}{*}{\shortstack[l]{FourRoom\\hard}} & NP & \priorstat{23.0}{4.2} & \priorstat{23.3}{3.7} & \priorstat{19.0}{2.7} \\
 & LP & \priorstat{56.0}{4.2} & \priorstat{70.0}{0.5} & \priorstat{100.0}{0.0} \\
 & $\Delta$P & +33.0 & +46.7 & +81.0 \\
\midrule
\multirow{3}{*}{\shortstack[l]{AntMaze\\OGBench}} & NP & \priorstat{0.0}{0.0} & \priorstat{55.3}{14.5} & \priorstat{44.0}{18.8} \\
 & LP & \priorstat{0.0}{0.0} & \priorstat{56.7}{11.5} & \priorstat{72.0}{10.2} \\
 & $\Delta$P & +0.0 & +1.4 & +28.0 \\
\bottomrule
\end{tabular*}
\endgroup
\end{table}

\begin{figure}[t]
\centering
\includegraphics[width=\columnwidth]{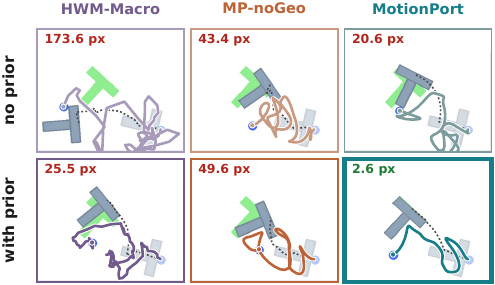}
\caption{\textbf{Effect of learned priors on PushT planning.} All methods share the same start and goal. Rows compare planning without (top) and with (bottom) learned priors. Only full \method{} with the prior successfully reaches the goal in this example.}
\label{fig:prior_refinement}\label{fig:module_interactions}
\end{figure}

\subsection{Learning Where to Search in Motion Space}
\label{sec:exp_proposals}

We next examine how learned priors improve planning in motion space (\cref{sec:motion-planning}; Appendix~D.2). As shown in \cref{tab:prior_closed_loop}, adding the prior increases \method{}'s success rate from 58.4\% to 97.3\% on Cube and from 19.0\% to 100\% on hard FourRoom. These gains are larger than those of MP-noGeo (28.2 and 46.7 percentage points, respectively), suggesting that learned initialization is more effective in a geometry-structured motion space. \Cref{fig:prior_refinement} illustrates this effect on PushT, where none of the methods reaches the goal without the prior, while only full \method{} succeeds with it. Together, these results suggest that the prior provides promising starting motions, while geometry makes it easier to refine them into actions that successfully reach the goal (Appendix~C.3).

\begin{figure}[t]
\centering
\includegraphics[width=.96\columnwidth]{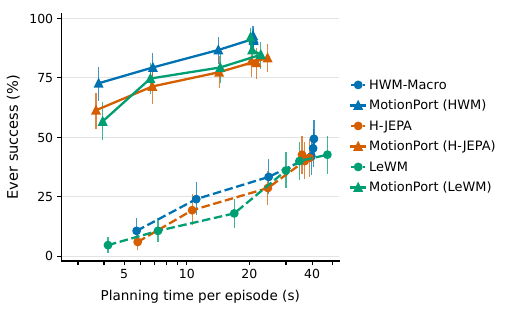}
\caption{\textbf{Planning efficiency across frozen world models.} Goal-reaching success versus planning time on long-horizon PushT. Colors distinguish LeWM, HWM, and H-JEPA, while dashed lines with circles represent native planners and solid lines with triangles represent \method{}. Using their frozen backbones, \method{} achieves higher success with substantially less planning time.}
\label{fig:planning_budget}

\par\medskip
\begingroup
\captionsetup{type=table,position=top}
\caption{\textbf{Motion-space planning better realizes predicted progress.} Both planners use the same temporal goal score on AntMaze-hard. PP (Predicted Progress) and RP (Realized Progress) measure reductions in $\phi$-distance based on predicted and actual outcomes. RPR (Realized-to-Predicted Ratio) is the ratio of mean RP to mean PP. NPE (Normalized Prediction Error) measures squared prediction error relative to squared observed state change.}
\label{tab:phi_delivery}
\centering\footnotesize
\setlength{\tabcolsep}{2pt}
\begin{tabular*}{\columnwidth}{@{\extracolsep{\fill}}lrrrr@{}}
\toprule
Planner & PP & RP & RPR & NPE \\
\midrule
HWM-Macro & 0.912 & 0.198 & 22\% & 2.68 \\
\method{} & 0.491 & \textbf{0.340} & \textbf{69\%} & \textbf{0.89} \\
\bottomrule
\end{tabular*}
\endgroup
\end{figure}

\begin{figure}[t]
\centering
\includegraphics[width=.96\columnwidth]{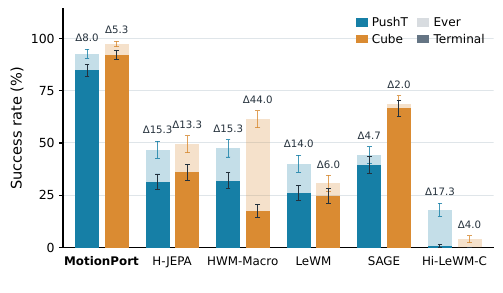}
\caption{\textbf{High goal-reaching success persists under continued execution.} Ever success measures whether a goal is reached at any point, while terminal success measures success at the final step. Results on PushT and Cube show that \method{} achieves both higher goal-reaching success and smaller gaps between ever and terminal success than most baselines.}
\label{fig:goal_retention}
\end{figure}

\subsection{Temporal Goal Geometry Guides Executable Plans}
\label{sec:exp_goal_geometry}

We next investigate whether a better goal score leads to better planning. As shown in \cref{fig:phi_score_landscape}(a--d), visual distance alone can favor states that are close to the goal but difficult to reach, while combining it with the learned temporal distance $\phi$ produces a score (\cref{eq:goal-cost}) that better reflects the maze's reachable paths. Across five Ant goals, the rank correlation with geodesic distance increases from 0.31 to 0.96 (Appendix~C.4). However, a better score does not benefit all planners equally. On AntMaze-hard, adding $\phi$ improves \method{}'s success rate from 38\% to 58\%, while HWM-Macro slightly decreases from 14\% to 12\%, as shown in \cref{fig:phi_score_landscape}(e).

To understand this difference, we compare the predicted and actual progress of the actions selected by each planner in \cref{tab:phi_delivery}. Although HWM-Macro predicts a larger reduction in goal distance (0.912 vs.\ 0.491), it achieves less progress after execution (0.198 vs.\ 0.340). Consequently, HWM-Macro realizes only 22\% of its predicted progress, compared with 69\% for \method{}, which also has a substantially lower prediction error (0.89 vs.\ 2.68).
This explains why improving the goal score alone is insufficient: HWM-Macro may favor actions whose predicted outcomes fail to materialize, whereas motion-space planning more reliably translates predicted progress into actual goal-reaching (Appendix~C.4).

\subsection{Efficient Planning across Frozen World Models}
\label{sec:exp_backbones}\label{sec:exp_budget}

We evaluate whether \method{} generalizes across different world models while maintaining planning efficiency. Averaged equally across 21 matched comparisons on LeWM, HWM, and H-JEPA, \method{} improves ever success from 37.9\% for native planning to 87.8\%, a gain of 50 percentage points, without fine-tuning the pretrained state encoders (Appendix~C.5). \Cref{fig:planning_budget} further compares their success rates under different planning budgets on long-horizon PushT (Appendices~C.6 and~D.4). Within a shared planning budget of 25 seconds per episode, \method{} outperforms native HWM, H-JEPA, and LeWM by 59.3, 54.7, and 74.0 percentage points, respectively. Notably, with HWM, \method{} achieves 72.7\% success in only about 4 seconds, exceeding the best native result of 49.3\% at approximately 40 seconds. Thus, \method{} achieves higher success while requiring only one-tenth of the planning time.

\subsection{Retaining a Goal under Continued Feedback}
\label{sec:exp_retention}

Finally, we evaluate whether goal-reaching success persists under continued closed-loop control. \Cref{fig:goal_retention} compares ever success with terminal success on PushT and Cube. \method{} achieves terminal success rates of 84.67\% and 92.00\%, respectively, substantially outperforming the strongest baselines at 39.33\% and 66.67\%. Moreover, its terminal success is only 8.0 and 5.3 percentage points below its ever success, indicating that most goals reached during an episode remain satisfied at the final step. 
These results show that \method{} not only reaches goals more reliably but also maintains high success under continued execution. Without reliable prediction of subsequent actions, continued replanning can move the object away from an already reached goal. MotionPort reduces this mismatch between predicted and actual outcomes, helping preserve goal success (Appendix~C.7).

\section{Conclusion}
We introduce \method{}, a geometry-structured motion interface that connects prediction, planning, and execution in frozen visual world models. By representing body motion rather than action chunks, \method{} provides a more effective optimization space, while learned priors and temporal goal scoring further improve planning. Experiments across manipulation and navigation tasks demonstrate substantial gains in goal-reaching success and planning efficiency across different world-model backbones. These results highlight the importance of action-space representation in bridging the gap between accurate prediction and effective control.

\label{end:maintext}

\clearpage

\section*{LLM Usage Statement}
We used OpenAI Codex and Anthropic Claude Code to assist with manuscript polishing and code development.

{\small\bibliographystyle{ieeenat_fullname}\bibliography{references}}
\clearpage
\setcounter{table}{0}
\setcounter{figure}{0}
\setcounter{equation}{0}
\renewcommand{\theHtable}{supp.\arabic{table}}
\renewcommand{\theHfigure}{supp.\arabic{figure}}
\renewcommand{\theHequation}{supp.\arabic{equation}}
\appendix
\setcounter{topnumber}{5}
\setcounter{bottomnumber}{3}
\setcounter{totalnumber}{8}
\setcounter{dbltopnumber}{5}
\renewcommand{\topfraction}{.92}
\renewcommand{\bottomfraction}{.85}
\renewcommand{\textfraction}{.06}
\renewcommand{\floatpagefraction}{.72}
\renewcommand{\dbltopfraction}{.92}
\renewcommand{\dblfloatpagefraction}{.72}
\setlength{\textfloatsep}{12pt plus 2pt minus 2pt}
\setlength{\dbltextfloatsep}{12pt plus 2pt minus 2pt}
\setlength{\floatsep}{10pt plus 2pt minus 2pt}
\setlength{\dblfloatsep}{10pt plus 2pt minus 2pt}
\raggedbottom
\makeatletter
\setlength{\@fptop}{0pt}
\setlength{\@fpsep}{12pt}
\setlength{\@fpbot}{0pt plus 1fil}
\setlength{\@dblfptop}{0pt}
\setlength{\@dblfpsep}{12pt}
\setlength{\@dblfpbot}{0pt plus 1fil}
\makeatother
\renewcommand{\thetable}{S\arabic{table}}
\renewcommand{\thefigure}{S\arabic{figure}}
\section{Evaluation Protocol and Motion Representation}
\label{sec:supp_protocol}

This supplement follows the organization of the main paper. Section~A
specifies the evaluation and motion-representation conventions;
Section~B explains the short- and long-goal comparison in Figure~1(c).
Sections~C.1--C.7 expand the prediction, optimization, prior, goal-score,
backbone, efficiency, and retention results. Section~D gives the training
and planning details of the main method.

\subsection{Tasks and reported success}
The seven evaluation settings use four environments. PushT and Cube
retain the original task definitions. FourRoom standard requires two
doorway crossings, while FourRoom hard requires three. AntMaze standard
uses the D3 goal set; AntMaze hard mixes D3 and longer D6 goals in the
New100 set. AntMaze OGBench uses the official long-range goals and the
stricter success threshold. The environment-step budget is shared across
methods within each setting.

\begin{table*}[!tbp]
\centering\small
\setlength{\tabcolsep}{5pt}
\begin{tabular*}{\textwidth}{@{\extracolsep{\fill}}p{.25\textwidth}rp{.57\textwidth}@{}}
\toprule
Setting & \shortstack{Evaluation\\cases} & Goal and success convention \\
\midrule
PushT & 150 & Public50 + Fresh100; joint pusher/block position and block angle \\
Cube & 150 & Public50 + Fresh100; cube position within 4\,cm of the target \\
FourRoom standard & 50 & Two doorway crossings; original environment success rule \\
FourRoom hard & 100 & Three doorway crossings; same rule and budget as standard \\
AntMaze standard & 50 & D3 goals; 1\,m success radius \\
AntMaze hard & 100 & New D3/D6 goals; 1\,m success radius \\
AntMaze OGBench & 25 & Five official goals with five initializations; 0.5\,m radius \\
\bottomrule
\end{tabular*}
\caption{\textbf{Seven evaluation settings.} All methods in a setting
receive the same evaluation goals and environment-step budget.}
\label{tab:supp_eval_settings}
\end{table*}

Ever success indicates that the task criterion is satisfied at least
once during an episode. Terminal success tests the final environment
step. Their difference measures the fraction of episodes that reach the
goal but do not finish there; it is not a test of uninterrupted success
after first arrival. PushT success requires a joint four-coordinate pusher/block position
error below 20 pixels and a wrapped block-angle error below $\pi/9$. Its object-only pixel distance,
used to annotate illustrations, is a diagnostic rather than a replacement
for the task's success predicate. Cube success depends on cube position,
not gripper position or object orientation.

\subsection{Training information and motion features}
The pretrained visual encoder remains frozen. For hierarchical backbones,
features are taken at the active planning level. All trainable components
use the offline training trajectories; goal relabeling draws goals from
those trajectories rather than the evaluation tasks.
The native LeWM interface uses primitive command blocks, whereas HWM and
H-JEPA use their learned macro-action interfaces.

The fixed extractor $X$ maps an observation to representative body
features. These describe the agent's own movement: pusher position in
PushT, end-effector pose and gripper configuration in Cube, realized agent
motion in FourRoom, and locomotion coordinates in AntMaze. A motion code
encodes a short trajectory of these features. It does not encode a task
goal or an object's desired future pose. The current body state $o$ can
contain additional information required to decode the motion, and is not
assumed to equal the representative feature $x$.

Motion features are scaled using training statistics before motion
reconstruction and distance alignment. The deterministic planning code
is the standardized encoder mean. MP-noGeo uses the same motion-code
dimension and learning formulation with the geometry weight set to zero.
Its decoder and predictor are trained for the corresponding motion
coordinates. In the main geometry loss, PushT, FourRoom, and AntMaze use
squared distances ($p=2$), normalized by their batch means; Cube uses
unsquared distances ($p=1$), normalized by their batch maxima. Gradients
are retained through the code-distance normalization factor. Dynamics targets stay in the coordinates of the frozen
visual encoder; the same encoder dimension alone does not make features
from different backbones interchangeable.

\subsection{Metrics and uncertainty}
Prediction metrics compare each forecast with the corresponding real
future observation. Latent SSE is meaningful within one feature space;
centimetre errors use the separately described position readout. The
ratio PE/VE compares planning and validation errors at the stated forecast
horizons. Ratios and percentage-point differences are computed before
rounding when unrounded estimates are available.

Parenthesized uncertainties are the reported standard errors over
evaluation units. Repeated calls within an episode remain together when
resampling prediction diagnostics. The OGBench analysis groups its five
initializations under each official goal. Intervals are explicitly marked
as confidence intervals where provided.

\subsection{Distances in the action representations}
We compare code distances with body-motion distances on 2,000 held-out
windows and 200,000 fixed pairs. These measurements test what the
coordinates represent. They are distinct from validation prediction
error and from the optimization slices in Section~C.2.
\begin{table}[!tbp]
\centering
\small
\setlength{\tabcolsep}{4pt}
\begin{tabular}{lrrr}
\toprule
Domain & \shortstack{Native\\input} & MP-noGeo & MotionPort \\
\midrule
PushT & 0.840 & \textbf{0.943} & 0.920 \\
Cube & 0.587 & 0.780 & \textbf{0.938} \\
FourRoom & 0.520 & 0.798 & \textbf{0.995} \\
Ant & 0.087 & 0.863 & \textbf{0.958} \\
\bottomrule
\end{tabular}
\caption{\textbf{Motion-distance alignment with LeWM.} Spearman correlation with body-motion distance on the same 2,000 held-out windows and 200,000 fixed pairs. Native input is the native primitive command chunk. The two motion variants both use four dimensions and share their representation across backbone instantiations. LeWM has no learned native macro-action; its input is the standardized command sequence. The native comparison changes the representation type; the two motion columns isolate geometry at fixed code dimension.}
\label{tab:supp_coordinates_lewm}
\end{table}
\begin{table}[!tbp]
\centering
\small
\setlength{\tabcolsep}{4pt}
\begin{tabular}{lrrr}
\toprule
Domain & HWM-Macro & MP-noGeo & MotionPort \\
\midrule
PushT & \textbf{0.943} & 0.943 & 0.920 \\
Cube & 0.286 & 0.780 & \textbf{0.938} \\
FourRoom & 0.488 & 0.798 & \textbf{0.995} \\
Ant & 0.591 & 0.863 & \textbf{0.958} \\
\bottomrule
\end{tabular}
\caption{\textbf{Motion-distance alignment with HWM.} Spearman correlation with body-motion distance on the same 2,000 held-out windows and 200,000 fixed pairs. Native input is the native learned macro-action. The two motion variants both use four dimensions and share their representation across backbone instantiations. Native macro-actions have four dimensions except Cube, where two four-dimensional macros cover the same twenty-step skill. The native comparison changes the representation type; the two motion columns isolate geometry at fixed code dimension.}
\label{tab:supp_coordinates_hwm}
\end{table}
\begin{table}[!tbp]
\centering
\small
\setlength{\tabcolsep}{4pt}
\begin{tabular}{lrrr}
\toprule
Domain & \shortstack{Native\\input} & MP-noGeo & MotionPort \\
\midrule
PushT & 0.942 & \textbf{0.943} & 0.920 \\
Cube & 0.279 & 0.780 & \textbf{0.938} \\
FourRoom & 0.488 & 0.798 & \textbf{0.995} \\
Ant & 0.609 & 0.863 & \textbf{0.958} \\
\bottomrule
\end{tabular}
\caption{\textbf{Motion-distance alignment with H-JEPA.} Spearman correlation with body-motion distance on the same 2,000 held-out windows and 200,000 fixed pairs. Native input is the native learned macro-action. The two motion variants both use four dimensions and share their representation across backbone instantiations. Native macro-actions have four dimensions except Cube, where two four-dimensional macros cover the same twenty-step skill. The native comparison changes the representation type; the two motion columns isolate geometry at fixed code dimension.}
\label{tab:supp_coordinates_hjepa}
\end{table}
Motion codes align body-motion distances across Cube, FourRoom, and
AntMaze. Section~C.2 examines how this structure shapes optimization
and physical execution.

\section{Short- and Long-Horizon PushT Evaluation}
\label{sec:supp_horizon}
This section documents main Figure~1(c). Short goals lie 25 logged steps ahead and use a 50-step control budget. The long-goal asset has a nominal offset of 75 frames (74 state transitions) and uses a 90-step budget. The two columns therefore compare their stated goal-and-budget protocols; they are not an equal-budget truncation experiment.

\begin{table*}[!tbp]
\centering\small
\begingroup
\newcommand{\horizonstat}[2]{#1\,{\scriptsize$(\pm #2)$}}
\setlength{\tabcolsep}{7pt}
\begin{tabular}{lrr}
\toprule
Method & Short-term ever success (\%) & Long-term ever success (\%) \\
\midrule
MotionPort & \horizonstat{93.33}{3.33} & \horizonstat{92.67}{2.13} \\
HWM & \horizonstat{44.89}{2.64} & \horizonstat{47.33}{4.07} \\
H-JEPA & \horizonstat{42.00}{2.77} & \horizonstat{46.67}{4.10} \\
SAGE & \horizonstat{43.78}{2.88} & \horizonstat{44.00}{4.07} \\
AP-CEM & \horizonstat{73.33}{3.62} & \horizonstat{41.33}{4.03} \\
LeWM & \horizonstat{81.78}{2.22} & \horizonstat{40.00}{4.03} \\
GC-IDM & \horizonstat{60.67}{4.00} & \horizonstat{30.67}{3.78} \\
Hi-LeWM-C & \horizonstat{23.56}{2.22} & \horizonstat{18.00}{3.15} \\
LeFlow & \horizonstat{81.56}{2.22} & \horizonstat{17.33}{3.11} \\
DINO-WM & \horizonstat{80.67}{3.24} & \horizonstat{10.67}{2.53} \\
Sub-JEPA & \horizonstat{84.67}{2.95} & \horizonstat{6.00}{1.95} \\
Qantara & \horizonstat{78.00}{2.63} & \horizonstat{4.89}{1.24} \\
Fast-LeWM & \horizonstat{82.67}{3.10} & \horizonstat{4.00}{1.61} \\
LP2 & \horizonstat{34.44}{3.52} & \horizonstat{2.67}{1.07} \\
\bottomrule
\end{tabular}
\endgroup
\caption{\textbf{Numerical values in the introductory comparison.} Means and error bars reproduce main Figure~1(c). Long-term values for the methods in main Table~1 agree with that table. The two horizons retain their respective goal and environment-step budgets.}
\label{tab:supp_short_long}
\end{table*}

Temporal abstraction also changes the planning schedule. In the native hierarchy, an upper layer is skipped when its remaining horizon reaches one, after which the lower layer targets the final goal directly. Under the standard short and long protocols, this transition occurs at steps 40 and 70, respectively. This explains why a longer-goal protocol need not have a lower observed success rate: it also allows more environment steps and changes when direct-goal optimization begins.

A separate Public50 intervention changes the native upper horizon to three while keeping its model weights fixed. The control below concerns planning timing, rather than the geometry intervention of MotionPort.

\begin{table}[t]
\centering\small
\begin{tabular}{lrrr}
\toprule
Method & Default short & Horizon-three short & Gain (pp) \\
\midrule
HWM & 40.00 & 70.00 & +30.00 \\
H-JEPA & 38.00 & 70.00 & +32.00 \\
\bottomrule
\end{tabular}
\caption{\textbf{A planning-schedule intervention on Public50.} Ever success (\%). This changes the prediction span, transition to direct-goal control, and subsequent feedback schedule together. It is a separate 50-task control.}
\label{tab:supp_horizon_control}
\end{table}

With the shorter upper horizon, the transition to direct-goal control occurs at step 30 and median first arrival moves to step 35. These results show that planning performance is sensitive to the horizon schedule.

\section{Additional Experimental Results and Analysis}
\label{sec:supp_planning_interface}
\raggedbottom
We first examine prediction accuracy on logged and planner-selected
motions, then relate code geometry to physical execution. The following
analyses study how priors and temporal goal scores guide search, how
MotionPort transfers across frozen backbones, and how planning efficiency
and goal retention change under the resulting controller.

\subsection{Prediction errors across all seven setups}
\label{sec:supp_prediction_complete}

\begin{table*}[!tbp]
\centering\small
\begingroup
\setlength{\tabcolsep}{3pt}
\renewcommand{\arraystretch}{1.10}
\newcommand{\predstat}[2]{#1\,{\normalfont\scriptsize$(\pm #2)$}}
\begin{tabular*}{\textwidth}{@{\extracolsep{\fill}}llrrr@{}}
\toprule
Setting & Metric & HWM-Macro & MP-noGeo & MotionPort \\
\midrule
\multirow{4}{*}{PushT} & VE & \predstat{2.81}{0.42} & \predstat{2.80}{0.48} & \predstat{2.89}{0.49} \\
 & PE & \predstat{19.00}{1.45} & \predstat{3.39}{0.23} & \predstat{3.43}{0.25} \\
 & ER & \predstat{6.76}{1.13} & \predstat{1.21}{0.22} & \predstat{1.19}{0.22} \\
 & SR & \predstat{47.33}{4.07} & \predstat{71.33}{2.31} & \predstat{92.67}{2.13} \\
\midrule
\multirow{4}{*}{Cube} & VE & \predstat{0.42}{0.01} & \predstat{0.70}{0.01} & \predstat{0.68}{0.01} \\
 & PE & \predstat{5.69}{0.19} & \predstat{1.38}{0.06} & \predstat{1.07}{0.04} \\
 & ER & \predstat{13.65}{0.53} & \predstat{1.97}{0.09} & \predstat{1.59}{0.06} \\
 & SR & \predstat{61.33}{4.00} & \predstat{85.33}{1.73} & \predstat{97.33}{1.31} \\
\midrule
\multirow{4}{*}{\shortstack[l]{FourRoom\\standard}} & VE & \predstat{0.49}{0.04} & \predstat{1.21}{0.09} & \predstat{1.41}{0.10} \\
 & PE & \predstat{2.86}{0.24} & \predstat{2.51}{0.09} & \predstat{2.17}{0.11} \\
 & ER & \predstat{5.80}{0.70} & \predstat{2.06}{0.16} & \predstat{1.55}{0.13} \\
 & SR & \predstat{98.00}{2.00} & \predstat{100.00}{0.00} & \predstat{100.00}{0.00} \\
\midrule
\multirow{4}{*}{\shortstack[l]{FourRoom\\hard}} & VE & \predstat{0.49}{0.04} & \predstat{1.21}{0.09} & \predstat{1.41}{0.10} \\
 & PE & \predstat{10.59}{0.72} & \predstat{2.62}{0.07} & \predstat{2.04}{0.06} \\
 & ER & \predstat{21.49}{2.36} & \predstat{2.16}{0.16} & \predstat{1.45}{0.12} \\
 & SR & \predstat{23.00}{4.23} & \predstat{70.0}{0.5} & \predstat{100.00}{0.00} \\
\midrule
\multirow{4}{*}{\shortstack[l]{AntMaze\\standard}} & VE & \predstat{3.30}{0.37} & \predstat{1.21}{0.25} & \predstat{1.17}{0.22} \\
 & PE & \predstat{30.21}{2.81} & \predstat{12.79}{1.36} & \predstat{11.27}{0.95} \\
 & ER & \predstat{9.16}{1.33} & \predstat{10.54}{2.55} & \predstat{9.64}{2.04} \\
 & SR & \predstat{60.00}{7.00} & \predstat{82.00}{5.49} & \predstat{92.00}{3.88} \\
\midrule
\multirow{4}{*}{\shortstack[l]{AntMaze\\hard}} & VE & \predstat{3.30}{0.37} & \predstat{1.21}{0.25} & \predstat{1.17}{0.22} \\
 & PE & \predstat{39.57}{1.86} & \predstat{14.88}{1.10} & \predstat{18.61}{1.43} \\
 & ER & \predstat{12.00}{1.46} & \predstat{12.26}{2.76} & \predstat{15.91}{3.30} \\
 & SR & \predstat{14.00}{4.29} & \predstat{63.00}{4.85} & \predstat{58.00}{4.96} \\
\midrule
\multirow{4}{*}{\shortstack[l]{AntMaze\\OGBench}} & VE & \predstat{3.27}{0.36} & \predstat{1.22}{0.24} & \predstat{1.18}{0.22} \\
 & PE & \predstat{38.31}{4.10} & \predstat{16.37}{3.81} & \predstat{15.73}{1.58} \\
 & ER & \predstat{11.71}{1.74} & \predstat{13.42}{4.02} & \predstat{13.37}{2.83} \\
 & SR & \predstat{0.00}{0.00} & \predstat{56.7}{11.5} & \predstat{72.00}{10.20} \\
\bottomrule
\end{tabular*}
\endgroup
\caption{\textbf{Validation and planning across all seven settings.}
VE is prediction error on held-out logged trajectories, PE is prediction
error on planner-selected commands, ER is PE/VE, and SR is ever success
(\%). Entries are mean (\(\pm\) SE). The PushT and Cube rows reproduce
main-paper Table~2; overlapping success results follow Tables~1 and~3.
Cube errors are three-dimensional position errors in cm through a common
diagnostic readout; other domains use latent SSE. Errors are compared
within a domain, in the frozen HWM feature coordinates. Evaluation tasks,
not individual replanning calls, are the sampling units; OGBench retains
its official-goal grouping. Ratios use unrounded error estimates.}
\label{tab:supp_prediction_all_settings}
\end{table*}

\paragraph{A common prediction comparison.}
\Cref{tab:supp_prediction_all_settings} extends
main-paper Table~2 to all seven evaluation settings, including the motion representation trained without
geometry. Native $F_1$ and each motion variant's own $G$ forecast the
outcome of their respective planner-selected commands. Validation uses
the same predictor and forecast lengths on logged motions, with lag
weights matched to each planning protocol. Errors use one common frozen
world-model coordinate system within each domain. Cube instead reports
3-D position error through its shared diagnostic readout; its centimetre
values should not be compared numerically with the latent SSE in other
domains.

\paragraph{Averaging over decisions.}
Each forecast is compared with the observation after its executed prefix.
Errors are averaged over the available transitions of an episode, and
then over evaluation episodes. Validation uses the analogous averaging
of held-out windows within their source episodes. ER is the ratio of the
two aggregate errors. The image used to start each new planning call is
observed, so errors from different feedback calls are not treated as one
uninterrupted imagined rollout.

\paragraph{Position readout used only for evaluation.}
The planner predicts visual features rather than decoded images. For Cube,
a common frozen diagnostic readout $R_{\rm pos}$ maps a forecast feature
to a three-dimensional cube position. It is fit on separate logged states
and shared by the compared methods. The error is
$100\|R_{\rm pos}(\hat h)-p_{\rm actual}\|_2$ in cm. Simulator cube
positions provide the evaluation target; this readout and target are not
inputs to the policy or its goal-scoring objective. Linear and MLP
readouts provide the robustness comparison in Section~C.7.

\paragraph{Validation and transition coverage.}
PushT uses all 21 episodes of the separate validation file, with four
sampled windows per episode. Its planning comparison retains 1,350
paired transitions over all 150 evaluation episodes: nine decisions
per episode, including the initial decision. The final call lacks the
matching next-observation error record used by this comparison.
HWM's upper encoder is the identity for this comparison, so $H_1$ and
$H_2$ are identical for all three predictors.

For FourRoom, validation uses 200 independent episodes from the original
collection procedure, yielding 800 held-out windows. Its validation errors are 0.49, 1.21, and 1.41
for native, no-geometry, and full-motion prediction. Ant uses 500 held-out episodes and 2,000 windows,
weighted by the executed forecast lengths of each goal set.

\paragraph{Interpreting the planning gap.}
Action support explains part of the other domains' differences. In
PushT, 4.1\% of native decisions leave the logged action range; native
PE is 94.34 on these decisions and 16.20 within the range, still well
above VE of 2.81. In FourRoom, the environment clamps commands to
$[-1,1]$. Predicting with the actions actually executed reduces native
PE to 0.48 on standard tasks and 1.40 on hard tasks, compared with
VE of 0.49. These checks distinguish the predictor's behavior on
executed actions from discrepancies between the commands it imagines
and those the environment executes.

\paragraph{Cube decision-index breakdown.}
For Cube, the main-paper prediction table includes every planning decision: four
per episode, at start steps 0, 10, 20, and 25. The aggregate includes the initial decision. Each
decision index and the three post-initial calls are also reported
separately to show how error changes over an episode.
All forecasts start from the current observed state. Consequently, a
larger error at a later decision is not, by itself, evidence of latent
rollout error accumulating across replanning calls. The native schedule
also changes: level 2 is solved at the initial call and skipped on the
subsequent calls.

\begin{table*}[!tbp]
\centering\small
\setlength{\tabcolsep}{7pt}
\begin{tabular}{lrrr}
\toprule
Cube position error (cm) & HWM-Macro $F_1$ & MP-noGeo & MotionPort \\
\midrule
Logged validation, $+5$ & 0.384 & 0.632 & 0.634 \\
Logged validation, $+10$ & 0.450 & 0.767 & 0.718 \\
\midrule
Planning, $0\to10$ & 1.37 & 0.33 & 0.49 \\
Planning, $10\to20$ & 6.55 & 1.67 & 1.02 \\
Planning, $20\to25$ & 7.52 & 1.47 & 1.48 \\
Planning, $25\to30$ & 7.34 & 2.05 & 1.30 \\
All four decisions (600 per system) & 5.69 & 1.38 & 1.07 \\
Post-initial subgroup (450 per system) & 7.14 & 1.73 & 1.27 \\
\bottomrule
\end{tabular}
\caption{\textbf{Prediction error by planning decision.} Results average over
150 Public50+Fresh100 episodes per HWM system. Position errors use a shared
MLP diagnostic readout of the imagined feature and the simulator's actual
cube position. Section~C.7 analyzes goal retention.}
\label{tab:supp_prediction_decisions}
\end{table*}

\begin{figure*}[!tbp]
\centering
\includegraphics[width=\textwidth,height=.60\textheight,keepaspectratio]{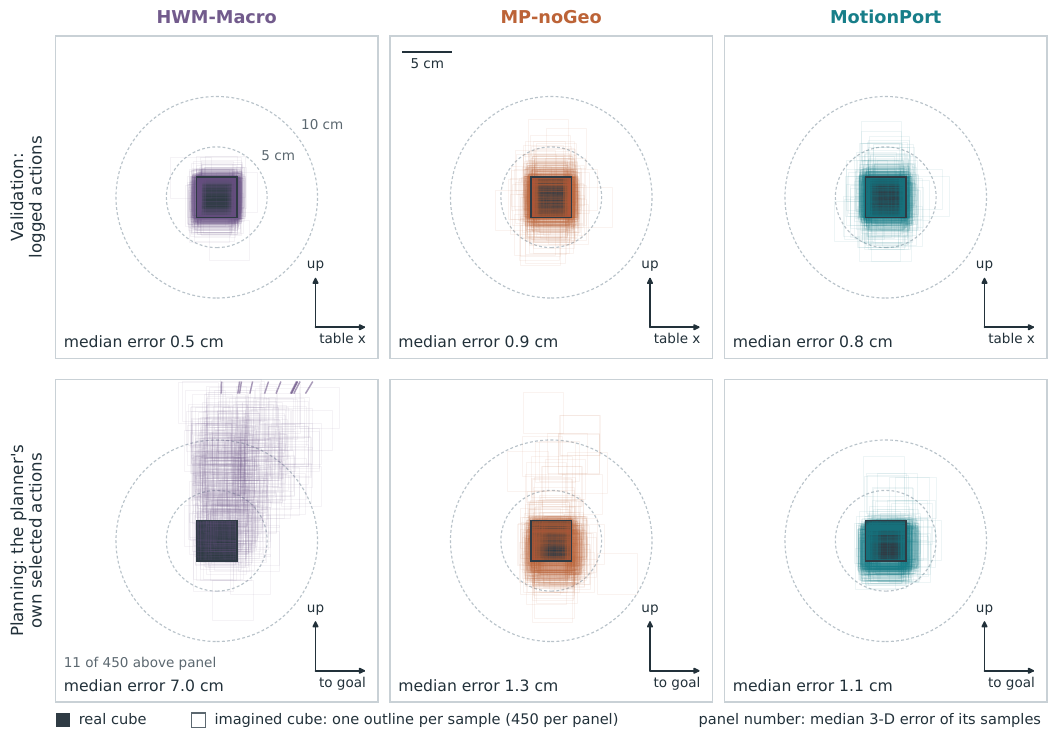}
\caption{\textbf{Where validation looks, and where each planner goes.}
Cube with frozen HWM. Each sample is translated so that its real cube
lies at the central filled square; each outline shows the predicted
position relative to that outcome, using a shared diagnostic readout.
Top: 450 sampled moving-cube validation windows at $+10$ steps;
``table x'' and ``up'' denote world-$x$ and height errors. Bottom: all
450 post-initial decisions across the same 150 tasks per system, pooling
executed 5- and 10-step blocks. Horizontal errors are rotated so that
``to goal'' points from the real cube toward the target in the table
plane; ``up'' remains height error. For horizontal goal distances below
0.5\,cm, the horizontal axis is world-$x$ (41 native,
111 no-geometry, and 143 full-motion samples). Thus the panels overlay prediction
errors, rather than absolute trajectories. Labels report median 3-D
position error; \cref{tab:supp_prediction_decisions} reports means. Native
forecasts are the final $F_1$ predictions after low-level optimization;
motion forecasts use the matching prefix of $G$. Later states and
commands follow each system's own policy.}
\label{fig:supp_planning_prediction}
\end{figure*}

\begin{table*}[!tbp]
\centering\small
\begin{tabular}{lrrr}
\toprule
Command source & HWM-Macro & MP-noGeo & MotionPort \\
\midrule
Same $F_1$, $+10$ feature SSE & 52.71 & 1.58 & 1.59 \\
95\% task-bootstrap interval & $[47.95,57.34]$ & $[1.24,1.97]$ & $[1.28,1.92]$ \\
\bottomrule
\end{tabular}
\caption{\textbf{Fixing the predictor and the current state.}
At the same 100 shared simulator states (two decisions on each of
50 development tasks), the same frozen $F_1$ predicts the outcomes of
candidate commands from each planning interface. Feature SSE compares
its forecast with the encoded image after actual 10-step execution;
no position readout is involved. Candidate pools have different
construction rules (35 native candidates and 40 motion candidates per
state), so this measures the respective queried action populations.
This is a separate shared-state cohort, not the 150-goal closed-loop
sample in \cref{tab:supp_prediction_decisions}.}
\label{tab:supp_same_f1}
\end{table*}

The fixed-predictor control first decodes motion codes to raw command
chunks $A=D(o,z)$, then passes $A$ to the unchanged native $F_1$ with
its native preprocessing. It changes the command source, not the
predictor. This differs from the full-system comparison, in which
native $F_1$ and motion-conditioned $G$ predict their respective
executed commands.

Table~2 of the main paper reports standard errors over evaluation units. Cube VE uses the 200
validation episode means, pairing the $+5$ and $+10$ windows within
each episode; PE uses all four decisions in each of 150 episodes,
stratified into Public50 and Fresh100. ER uses the standard deviation
of 10,000 bootstrap ratios, independently resampling validation
episodes and stratified planning tasks. AntMaze OGBench PE and SR use five official-goal cluster means. These
uncertainties describe evaluation variation at fixed model weights.

In the native population at decision 1, the mean $F_1$ feature SSE is
31.47 for initial random candidates and 89.70 for the optimized top-20.
This increase shows that optimization can select commands with larger
forecast errors. Motion-code candidates keep the same native predictor
accurate, linking the planning representation to the reliability of the
actions queried during search.

The consequences of these prediction errors after reaching a goal are
analyzed separately in \cref{sec:supp_retention_complete}, including
all decision counts and the common position-readout checks.

\subsection{Code distributions and execution maps}
\label{sec:supp_geometry}

\paragraph{Optimization and execution on a common slice.}
For each planning interface, we form a two-dimensional affine slice
through the initial code $z^{(0)}$. Its first basis vector follows the
initial-to-final displacement of a 30-update optimization trajectory;
the second captures the largest orthogonal excursion of that trajectory.
A point $(u,v)$ therefore denotes $z^{(0)}+u e_1+v e_2$ in the original
planning coordinates.

We evaluate predicted costs on a common displayed rectangle using a
$51\times51$ grid. Costs are normalized by $J_{\rm ref}$, the objective
at the initial reference code. Starting-point selection targets a comparable normalized cost of
$J/J_{\rm ref}=2$, followed by 30 AdamW updates within the slice. We mark the first update
reaching $J/J_{\rm ref}\leq0.6$ and the final iterate.

For the execution map, each grid code generates a command sequence
that is executed from the same physical state for 20 steps. Native
macro-actions use the original low-level solver; motion codes use their
decoders. The lower row plots cube-to-goal distance and its 4\,cm
success boundary along the same optimization path.

The native Cube macro-action encoder applies layer normalization to its
four outputs. Logged codes lie on the resulting sphere, while optimized
codes can leave it. With the same 2,000-code validation-reference size,
59\% of selected native codes exceed the validation 99th-percentile
nearest-neighbor spacing; the corresponding proportions are 0.3\% for
full MotionPort and 0.7\% without geometry. The execution maps then assess how these distribution shifts relate
to physical outcomes.

\begin{figure*}[!tbp]
\centering
\includegraphics[width=\textwidth,height=.43\textheight,keepaspectratio]{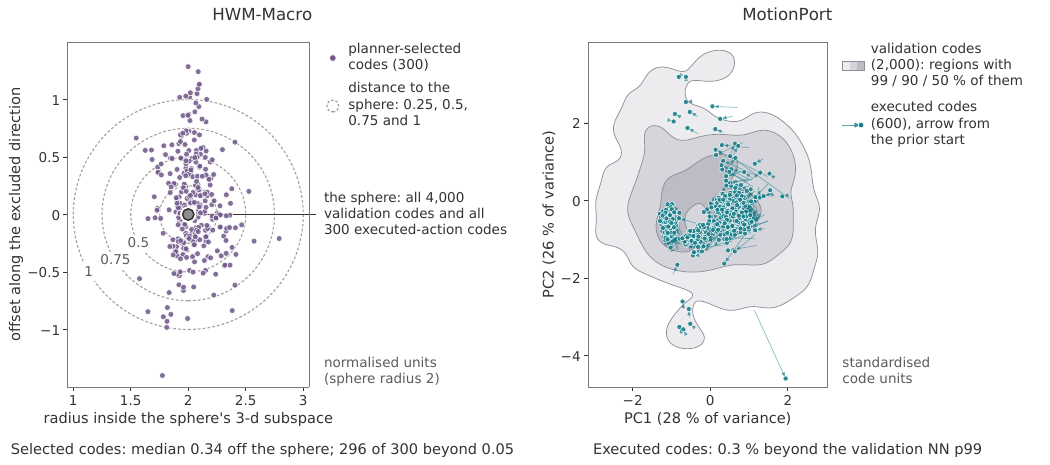}
\caption{\textbf{Validation codes and selected planning codes.}
Native macro coordinates and projected motion coordinates are shown
with their validation distributions. Nearest-neighbor statistics use
equally sized reference sets. Execution outcomes are evaluated in the
code-slice experiments.}
\label{fig:supp_code_spaces}
\end{figure*}

Across ten Cube development states, MotionPort reduces neighboring-code
success disagreement from 24.5\% to 2.6\% and the mean neighboring-code
goal-distance difference from 3.07 to 0.35\,cm. These measurements show
that nearby motion codes produce more consistent physical outcomes.

The experiment uses one planning decision from each of ten Cube
development tasks. Its native grids use $17\times17$
nodes and motion grids $26\times26$ nodes. At the single displayed state,
we additionally execute equal-resolution $26\times26$ grids and dense
$51\times51$ grids. Halving the spacing changes native neighbor success
disagreement from 45.5\% to 44.7\%, while MotionPort changes from 4.5\%
to 2.3\%. Mean neighbor distance differences change from 4.82 to
4.91\,cm for native macros and from 0.29 to 0.15\,cm for MotionPort.
The dependence on the low-level solver is checked separately by using a
common solver initialization across all native codes.

Across ten decisions, agreement of native-code success under the two
solver-initialization schemes is 76.4\%, compared with a chance agreement of
71.5\% (mean Cohen's $\kappa=0.17$). Execution therefore depends on both the macro-action code and the
lower-level solve.

\subsection{Learned-prior initialization and physical execution}
\label{sec:supp_prior}
\begin{table*}[!tbp]
\centering\small
\begingroup
\newcommand{\suppstat}[2]{#1\,{\scriptsize$(\pm #2)$}}
\setlength{\tabcolsep}{5pt}
\begin{tabular*}{\textwidth}{@{\extracolsep{\fill}}llrrr@{}}
\toprule
Setup & Interface & No prior & Learned prior & $\Delta$P (pp) \\
\midrule
PushT & HWM-Macro & \suppstat{47.3}{4.1} & \suppstat{55.8}{2.8} & +8.5 \\
 & MP-noGeo & \suppstat{62.6}{2.3} & \suppstat{71.3}{2.3} & +8.7 \\
 & MotionPort & \suppstat{80.9}{2.5} & \suppstat{92.7}{2.1} & +11.8 \\
\addlinespace
Cube & HWM-Macro & \suppstat{61.3}{4.0} & \suppstat{71.1}{2.5} & +9.8 \\
 & MP-noGeo & \suppstat{57.1}{2.3} & \suppstat{85.3}{1.7} & +28.2 \\
 & MotionPort & \suppstat{58.4}{2.7} & \suppstat{97.3}{1.3} & +38.9 \\
\addlinespace
FourRoom hard & HWM-Macro & \suppstat{23.0}{4.2} & \suppstat{56.0}{4.2} & +33.0 \\
 & MP-noGeo & \suppstat{23.3}{3.7} & \suppstat{70.0}{0.5} & +46.7 \\
 & MotionPort & \suppstat{19.0}{2.7} & \suppstat{100.0}{0.0} & +81.0 \\
\addlinespace
AntMaze OGBench & HWM-Macro & \suppstat{0.0}{0.0} & \suppstat{0.0}{0.0} & +0.0 \\
 & MP-noGeo & \suppstat{55.3}{14.5} & \suppstat{56.7}{11.5} & +1.4 \\
 & MotionPort & \suppstat{44.0}{18.8} & \suppstat{72.0}{10.2} & +28.0 \\
\addlinespace
\bottomrule
\end{tabular*}
\endgroup
\caption{\textbf{Prior effects in the main configurations.} Ever success (\%, mean $\pm$ SE). NP uses Gaussian initialization; LP allocates half of the initial proposals to the learned-prior mixture and half to Gaussian samples. The learned component combines the goal-conditioned and goal-free priors described in Appendix~D.2. This table expands the prior comparison in main-paper Table~3.}
\label{tab:supp_prior_summary}
\end{table*}

Without a learned prior, candidates are initialized from a Gaussian in
each interface's planning space. With the prior, half of the candidates
come from the learned goal-conditioned and goal-free models and half
from the Gaussian. One-motion sampling alternates with prediction to
initialize sequences; all codes are then optimized jointly.

The prior increases MotionPort's ever success by 38.9 percentage points
on Cube and 81.0 on FourRoom hard, compared with 28.2 and 46.7 for
MP-noGeo. We examine how initialization changes the candidate population
and how candidate refinement translates into executed actions.

\paragraph{Experimental protocol.}
The Cube population study evaluates 200 refined candidates for each
interface and prior condition at each of ten development states. The
code-space illustration uses one of these states and shows 400
candidates per condition: all motion candidates and a fixed half of the
native planner's 800 candidates. Native macro-actions are decoded using
independently initialized low-level solves; their population statistics
therefore include variability from this solver. The closed-loop study
uses 150 evaluation goals and the commands selected by each controller.
Both studies measure whether the gripper reaches within 3\,cm of the
cube after ten steps; closed-loop evaluation additionally measures
whether the cube reaches its task goal.

\begin{figure*}[!tbp]
\centering
\includegraphics[width=\textwidth]{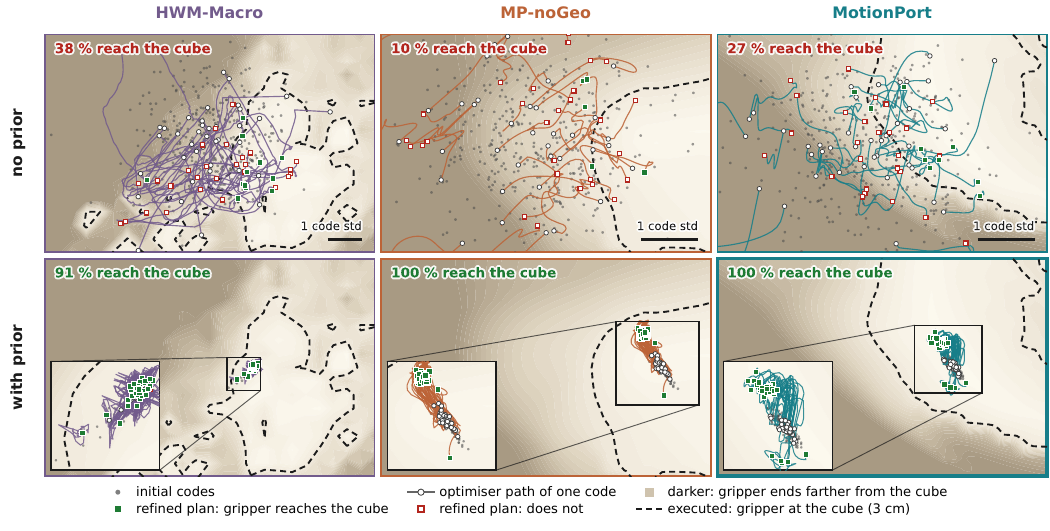}
\caption{\textbf{Where the prior places search in code space.}
Cube's first planning decision with frozen HWM. Rows compare Gaussian
initialization and learned priors; columns compare HWM-Macro, MP-noGeo,
and MotionPort. Dots and curves show initial codes and optimization
trajectories. Shading gives gripper-to-cube distance after ten steps;
the dashed contour marks 3\,cm. Green squares and percentages indicate
refined candidates meeting this approach criterion, among 400 candidates
per panel. The population protocol is described above; closed-loop
outcomes appear in \cref{tab:supp_prior_first_closed_loop}.}
\label{fig:supp_prior_code_space}
\end{figure*}

\paragraph{Code-space refinement.}
Each plane in \cref{fig:supp_prior_code_space} passes through the
no-prior mean, the prior's initial mean, and its mean refined code.
Points and trajectories are orthogonal projections of full-dimensional
candidates. Background colors result from executing grid codes in the
plane; candidate markers use executions of the full codes.

\begin{table*}[!tbp]
\centering\small
\setlength{\tabcolsep}{6pt}
\begin{tabular}{lrrr}
\toprule
Interface & No prior (\%) & With prior (\%) & Gain (pp) \\
\midrule
HWM-Macro & 63.15 $[51.15, 75.00]$ & 92.65 $[89.45, 95.65]$ & 29.50 $[19.45, 39.60]$ \\
MP-noGeo & 8.50 $[5.50, 11.65]$ & 99.80 $[99.40, 100.00]$ & 91.30 $[88.30, 94.15]$ \\
MotionPort & 21.15 $[18.05, 24.50]$ & 99.90 $[99.70, 100.00]$ & 78.75 $[75.40, 81.95]$ \\
\bottomrule
\end{tabular}
\caption{\textbf{Candidate approach rates across ten Cube states.}
Share of refined candidates bringing the gripper within 3\,cm of the
cube after ten steps. Each state contributes 200 candidates per
interface and prior condition. Brackets give paired 95\% bootstrap
intervals over states. This intermediate approach criterion is distinct
from task success.}
\label{tab:supp_prior_first_candidates}
\end{table*}

\begin{table*}[!tbp]
\centering\small
\begingroup
\newcommand{\firststat}[2]{#1\,{\scriptsize$(\pm #2)$}}
\setlength{\tabcolsep}{6pt}
\begin{tabular}{llrr}
\toprule
Interface & Prior & Gripper at cube at step 10 (\%) & Ever task success (\%) \\
\midrule
HWM-Macro & No prior & \firststat{75.8}{2.2} & \firststat{61.3}{4.0} \\
HWM-Macro & Learned prior & \firststat{91.1}{1.3} & \firststat{71.1}{2.5} \\
MP-noGeo & No prior & \firststat{23.1}{2.4} & \firststat{57.1}{2.3} \\
MP-noGeo & Learned prior & \firststat{98.9}{0.5} & \firststat{85.3}{1.7} \\
MotionPort & No prior & \firststat{41.3}{2.9} & \firststat{58.4}{2.7} \\
MotionPort & Learned prior & \firststat{98.4}{0.7} & \firststat{97.3}{1.3} \\
\bottomrule
\end{tabular}
\endgroup
\caption{\textbf{The same intermediate condition in closed-loop control.}
Cube, with the 150 evaluation goals. The intermediate condition tests
whether the gripper is within 3\,cm of the cube at step 10. Ever success
uses the official cube-to-goal criterion. Parentheses report SEs. The intermediate condition
and goal-reaching are different metrics; a successful approach does not
by itself establish successful manipulation.}
\label{tab:supp_prior_first_closed_loop}
\end{table*}

\paragraph{From candidates to executed actions.}
MotionPort checks the 20 lowest-cost candidates with the frozen native
predictor before execution. This re-ranking replaces the minimum-$J$
candidate in 71--90\% of the first planning calls in this study.

\begin{figure*}[!tbp]
\centering
\includegraphics[width=\textwidth]{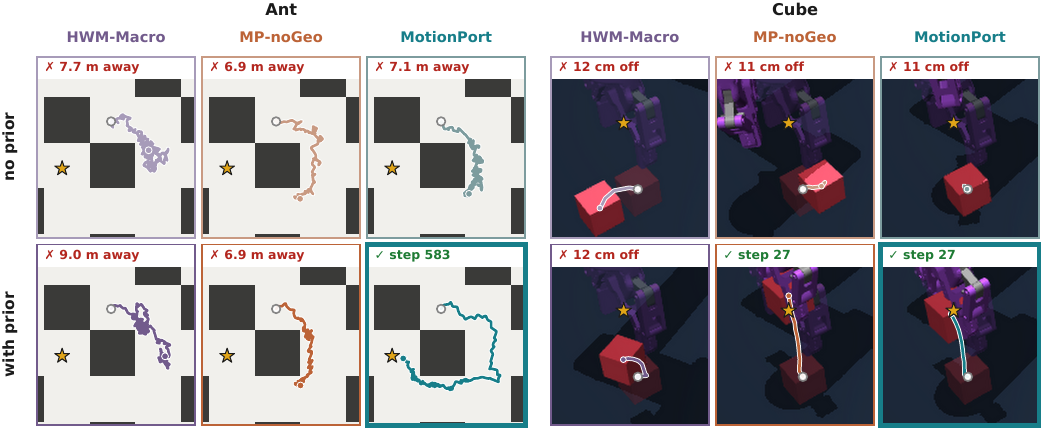}
\caption{\textbf{Additional prior comparisons.}
Rows compare Gaussian initialization and learned priors; columns compare
HWM-Macro, MP-noGeo, and MotionPort. All conditions share the start and
goal within each task. In the AntMaze OGBench example, only MotionPort with the learned prior
reaches the goal, at step 583; this pattern occurs in 3 of the 25 tasks.
In Cube, both motion variants with the prior reach the goal at step 27,
while the other four conditions fail. The Cube example illustrates the
prior's benefit for both motion representations. Scenes use the original
renderer, with Ant paths overlaid on the maze. Checks denote ever
success; failed cases show the final goal offset.}
\label{fig:supp_prior_scenes}
\end{figure*}

\paragraph{Rendered examples.}
Main Figure~5 and \cref{fig:supp_prior_scenes} use the environment's
renderer at physical states, rather than images reconstructed from
predicted features. PushT's pixel annotation measures block-position
error; its success mark uses the joint position-and-orientation rule in
Section~A. These examples illustrate refinement from a shared initial
state and goal, while the prior tables measure aggregate performance.

\subsection{Temporal scoring: spatial signal and delivered progress}
\label{sec:supp_goal_score}
We evaluate the temporal score used by HWM-based MotionPort. Spatial maps use 1,280 rendered Ant
positions and five goals. Mean geodesic rank correlation is 0.31 for
visual distance and 0.96 for the deployed mixed score. Geodesic distances are used only to evaluate the landscape; the temporal
embedding is learned from transitions, as specified in Section~D.3.

\begin{table*}[!tbp]
\centering\small
\setlength{\tabcolsep}{6pt}
\begin{tabular}{lrrrr}
\toprule
Goal & $\rho(E_2)$ & $\rho(E_2+\phi)$ & Misleading $E_2$ (\%) & Misleading mixed (\%) \\
\midrule
1 & 0.36 & 0.98 & 37.9 & 14.6 \\
2 & 0.47 & 0.96 & 34.0 & 16.4 \\
3 & 0.23 & 0.97 & 32.7 & 14.7 \\
4 & 0.16 & 0.96 & 34.5 & 16.4 \\
5 & 0.33 & 0.95 & 28.7 & 15.2 \\
\midrule
Mean & 0.31 & 0.96 & 33.6 & 15.5 \\
\bottomrule
\end{tabular}

\caption{\textbf{Spatial scoring statistics for every OGBench goal.}
The main paper displays goals 1 and 3. Here each row uses all 1,280
sampled Ant positions. $E_2$ is visual-feature distance; the mixed score
uses the deployed normalized weights 0.05 visual and 0.95 temporal.
$\rho$ compares the score ranking with geodesic maze distance.
A misleading location is one whose lowest-cost four-neighbor has lower
score but greater geodesic distance. These are deterministic spatial
diagnostics, not control success rates.}
\label{tab:supp_phi_landscape}
\end{table*}

\paragraph{Goal-reaching with visual and temporal scores.}
\Cref{tab:phi_native} compares native and motion planning on the three
AntMaze settings. \Cref{tab:supp_phi_complete_hwm} gives the motion-space
score ablation across all seven settings. The same head and normalization
are used for the native-versus-MotionPort intervention. The displayed
spatial maps use official OGBench goals, whereas main Figure~4(e) uses the
separate AntMaze-hard evaluation set; the map examples and the bar chart
therefore answer complementary questions.

\begin{table*}[!tbp]
\centering\small
\begingroup
\newcommand{\suppstat}[2]{#1\,{\scriptsize$(\pm #2)$}}
\begin{tabular*}{\textwidth}{@{\extracolsep{\fill}}llrrr@{}}
\toprule
Setup & Planner & $E_2$ & $E_2+\phi$ & Gain (pp) \\
\midrule
AntMaze standard & HWM-Macro & \suppstat{60.00}{7.00} & \suppstat{60.0}{7.0} & +0.00 \\
AntMaze standard & MotionPort & \suppstat{60.0}{7.0} & \suppstat{92.00}{3.88} & +32.00 \\
AntMaze hard & HWM-Macro & \suppstat{14.00}{4.29} & \suppstat{12.0}{3.3} & -2.00 \\
AntMaze hard & MotionPort & \suppstat{38.0}{4.9} & \suppstat{58.00}{4.96} & +20.00 \\
AntMaze OGBench & HWM-Macro & \suppstat{0.00}{0.00} & \suppstat{0.0}{0.0} & +0.00 \\
AntMaze OGBench & MotionPort & \suppstat{0.0}{0.0} & \suppstat{72.00}{10.20} & +72.00 \\
\bottomrule
\end{tabular*}
\endgroup
\caption{\textbf{The same temporal scoring rule in native and motion planning.} Ever success (\%, mean $\pm$ SE). Native visual references follow main-paper Table~1; the AntMaze-hard means reproduce Figure~4(e). The two methods use the same goal-scoring head, mixture, and normalization in this intervention. The standard and hard tasks use a 1\,m threshold; OGBench uses 0.5\,m.}
\label{tab:phi_native}
\end{table*}

\begin{table*}[!tbp]
\centering\small
\begingroup
\newcommand{\suppstat}[2]{#1\,{\scriptsize$(\pm #2)$}}
\setlength{\tabcolsep}{5pt}
\begin{tabular*}{\textwidth}{@{\extracolsep{\fill}}lrrr@{}}
\toprule
Setup & Visual score $E$ & Mixed score $E+\phi$ & Gain (pp) \\
\midrule
PushT & \suppstat{89.3}{2.5} & \suppstat{92.67}{2.13} & +3.37 \\
Cube & \suppstat{96.7}{1.5} & \suppstat{97.33}{1.31} & +0.63 \\
FourRoom standard & \suppstat{100.0}{0.0} & \suppstat{100.00}{0.00} & +0.00 \\
FourRoom hard & \suppstat{100.0}{0.0} & \suppstat{100.00}{0.00} & +0.00 \\
AntMaze standard & \suppstat{60.0}{7.0} & \suppstat{92.00}{3.88} & +32.00 \\
AntMaze hard & \suppstat{38.0}{4.9} & \suppstat{58.00}{4.96} & +20.00 \\
AntMaze OGBench & \suppstat{0.0}{0.0} & \suppstat{72.00}{10.20} & +72.00 \\
\bottomrule
\end{tabular*}
\endgroup
\caption{\textbf{MotionPort goal scoring across the seven HWM settings.}
Ever success (\%, mean $\pm$ SE). The mixed-score column matches
main-paper Table~1. $E$ uses visual distance and $E+\phi$ uses the
normalized visual--temporal score. Gains are percentage-point differences
of the displayed means. The corresponding native-versus-MotionPort
intervention is reported in \cref{tab:phi_native}.}
\label{tab:supp_phi_complete_hwm}
\end{table*}

OGBench uses the official 0.5\,m success radius; AntMaze standard and
hard use a 1\,m radius.

With the shared mixed score, native ever success on
Public50/New100/OGBench is 60/12/0\%, compared with 92/58/72\% for
MotionPort.

\begin{figure*}[!tbp]
\centering
\includegraphics[width=\textwidth,height=.55\textheight,keepaspectratio]{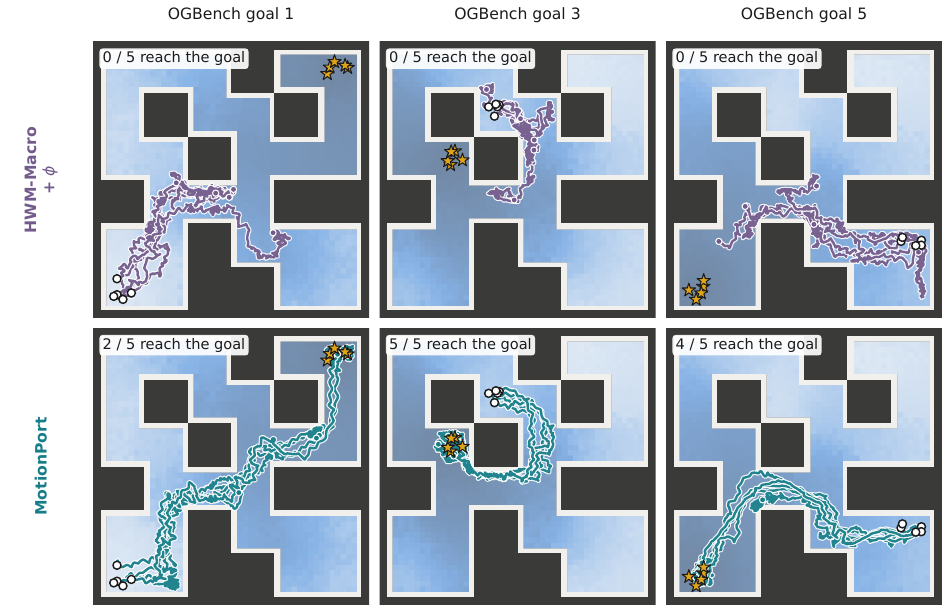}
\caption{\textbf{The same temporal score must be translated into real progress.}
Ant trajectories with frozen HWM, OGBench goals 1, 3, and 5, five
initializations per goal. Top: the native planner with the same mixed
goal score used by \method{}; bottom: \method{}. Background color shows
the shared mixed goal cost; white circles mark starts, and stars mark goals.
Across all five goals, the two systems reach 0/25 and 18/25, respectively.
}
\label{fig:phi_trajectories}
\end{figure*}

\begin{figure*}[!tbp]
\centering
\includegraphics[width=\textwidth,height=.36\textheight,keepaspectratio]{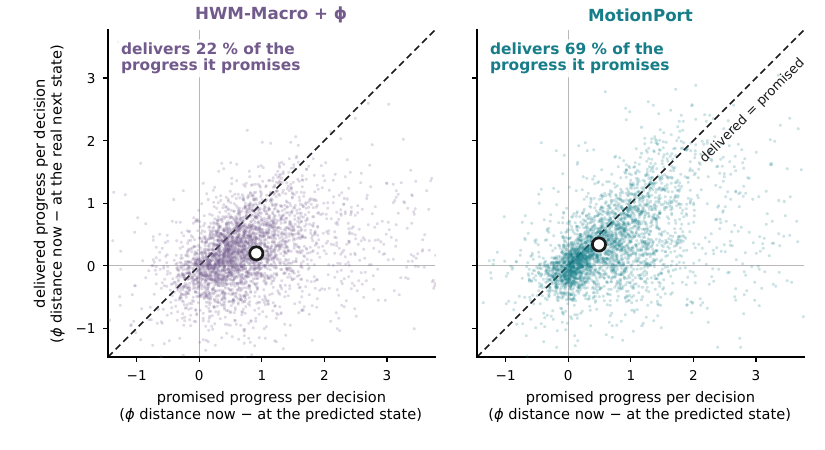}
\caption{\textbf{Promised and delivered temporal progress.}
Each point is one planning decision on the 100 new Ant goals; rings
show means. The decrease predicted by each planner is compared with
the decrease at its real next state. Under the same mixed score,
native planning delivers 22\% of its mean promised progress and
MotionPort delivers 69\%.}
\label{fig:supp_phi_progress}
\end{figure*}

\paragraph{Progress and prediction definitions.}
For each call, let $h$ be its observed feature, $\hat h^+$ the forecast
at the end of the prefix actually executed, and $h^+$ the next observed
feature. Promised and delivered progress are respectively
$d_\phi(h,h_g)-d_\phi(\hat h^+,h_g)$ and
$d_\phi(h,h_g)-d_\phi(h^+,h_g)$, evaluated with the same frozen head.
The displayed delivery fraction is the ratio of the two means, not the
mean of per-call ratios. Native forecasts come from the level-1 prediction
at the actual execution horizon; motion forecasts come from the matching
prefix head of $G$. The last call is omitted when its next encoded
observation is unavailable. The analysis includes every other call, including the first.  Normalized error is
$\sum\lVert\hat h^+-h^+\rVert^2 / \sum\lVert h-h^+\rVert^2$.

\begin{table*}[!tbp]
\centering\small
\setlength{\tabcolsep}{4pt}
\begin{tabular}{lrrr}
\toprule
New100 system & Promised progress & Delivered progress & Normalized prediction error \\
\midrule
HWM-Macro, visual score & 0.060 & 0.078 & 3.96 \\
HWM-Macro, shared mixed score & 0.912 & 0.198 & 2.68 \\
MotionPort, visual score & $-0.469$ & 0.158 & 4.62 \\
MotionPort, shared mixed score & 0.491 & 0.340 & 0.89 \\
\bottomrule
\end{tabular}
\caption{\textbf{Predicted and realized progress on AntMaze-hard.}
100 tasks and 3,500 decisions per system, frozen HWM. Normalized
error is next-state prediction error relative to the real state change;
values above one are worse than predicting no change. A negative mean
promise does not define a useful delivered/promised ratio.}
\label{tab:supp_phi_progress}
\end{table*}

Together, the progress and prediction results show how motion-space
planning translates temporal-score improvements into physical progress.

\subsection{Frozen-backbone comparisons}
\label{sec:supp_backbones}
\Cref{tab:supp_backbone_complete} expands the three-backbone comparison
in main-paper Section~4.5. MotionPort keeps the pretrained state encoder frozen and learns its
motion-conditioned predictor in the corresponding feature space. The
table reports ever success and the gain over native planning for each
backbone and evaluation setting.
\begin{table}[!htbp]
\centering\footnotesize
\setlength{\tabcolsep}{2pt}
\begin{tabular*}{\columnwidth}{@{\extracolsep{\fill}}llrrr@{}}
\toprule
Backbone & Setup & Native & MotionPort & Gain (pp) \\
\midrule
LeWM & PushT & 40.00 & 85.91 & +45.91 \\
LeWM & Cube & 30.67 & 99.21 & +68.54 \\
LeWM & FourRoom standard & 58.00 & 99.21 & +41.21 \\
LeWM & FourRoom hard & 0.00 & 99.21 & +99.21 \\
LeWM & AntMaze standard & 24.00 & 91.21 & +67.21 \\
LeWM & AntMaze hard & 3.00 & 68.21 & +65.21 \\
LeWM & AntMaze OGBench & 0.00 & 83.21 & +83.21 \\
\addlinespace
HWM & PushT & 47.33 & 92.67 & +45.33 \\
HWM & Cube & 61.33 & 97.33 & +36.00 \\
HWM & FourRoom standard & 98.00 & 100.00 & +2.00 \\
HWM & FourRoom hard & 23.00 & 100.00 & +77.00 \\
HWM & AntMaze standard & 60.00 & 92.00 & +32.00 \\
HWM & AntMaze hard & 14.00 & 58.00 & +44.00 \\
HWM & AntMaze OGBench & 0.00 & 72.00 & +72.00 \\
\addlinespace
H-JEPA & PushT & 46.67 & 81.21 & +34.54 \\
H-JEPA & Cube & 49.33 & 97.21 & +47.88 \\
H-JEPA & FourRoom standard & 98.00 & 99.21 & +1.21 \\
H-JEPA & FourRoom hard & 41.00 & 99.21 & +58.21 \\
H-JEPA & AntMaze standard & 78.00 & 91.21 & +13.21 \\
H-JEPA & AntMaze hard & 22.82 & 71.21 & +48.39 \\
H-JEPA & AntMaze OGBench & 0.00 & 67.21 & +67.21 \\
\addlinespace
\bottomrule
\end{tabular*}
\caption{\textbf{Goal-reaching across three frozen backbones.} Ever success (\%). Native references and all HWM MotionPort results reproduce main-paper Table~1. The remaining rows report the corresponding MotionPort backbone configurations. Gains are differences in percentage points; small discrepancies between displayed subtraction and a reported gain can result from rounding.}
\label{tab:supp_backbone_complete}\label{tab:frozen_backbones}\label{fig:frozen_backbones}
\end{table}

\subsection{Planning budgets and timing}
\label{sec:supp_timing}
The main timing comparison uses long-horizon PushT and the same
150-goal evaluation set for every point. \Cref{tab:supp_timing_points}
reports all 36 operating points shown in main Figure~6. Times include visual encoding
and planning and exclude simulator stepping. Measurements use an
otherwise idle L40S. Per-episode planning time accounts for the number
of planning calls; reusing a cached action is not a new optimization.
Each method's budget sweep varies its own candidate and optimizer budget,
while keeping its trained modules fixed.

\begin{table}[!htbp]
\centering\footnotesize
\setlength{\tabcolsep}{3pt}
\begin{tabular*}{\columnwidth}{@{\extracolsep{\fill}}llrr@{}}
\toprule
Backbone & Budget & Native & MotionPort \\
\midrule
HWM & 1/16 & 10.7 / 5.74 & 72.7 / 3.77 \\
 & 1/8 & 24.0 / 11.11 & 79.3 / 6.87 \\
 & 1/4 & 33.3 / 24.77 & 86.7 / 14.18 \\
 & 1/2 & 42.0 / 39.75 & 91.3 / 20.94 \\
 & 1 & 49.3 / 40.86 & 92.7 / 20.83 \\
 & 2 & 45.3 / 40.48 & 90.7 / 20.94 \\
\addlinespace
H-JEPA & 1/16 & 6.0 / 5.81 & 61.3 / 3.66 \\
 & 1/8 & 19.3 / 10.61 & 71.3 / 6.83 \\
 & 1/4 & 28.7 / 24.54 & 77.3 / 14.31 \\
 & 1/2 & 40.0 / 36.89 & 83.3 / 24.45 \\
 & 1 & 42.7 / 35.82 & 82.0 / 20.40 \\
 & 2 & 41.3 / 38.87 & 81.3 / 21.61 \\
\addlinespace
LeWM & 1/16 & 4.7 / 4.18 & 56.7 / 3.94 \\
 & 1/8 & 10.7 / 7.27 & 74.7 / 6.68 \\
 & 1/4 & 18.0 / 16.91 & 79.3 / 14.48 \\
 & 1/2 & 36.0 / 29.97 & 84.7 / 22.70 \\
 & 1 & 40.0 / 34.82 & 86.7 / 20.61 \\
 & 2 & 42.7 / 47.44 & 92.0 / 20.25 \\
\addlinespace
\bottomrule
\end{tabular*}
\caption{\textbf{All operating points in main Figure~6.} Each cell is ever success (\%) / planning seconds per episode. Budget is the method-specific compute multiplier, not an environment-step budget. The success means and timing coordinates reproduce the main figure; its error bars are shown there.}
\label{tab:supp_timing_points}
\end{table}

\begin{table}[t]
\centering\small
\begin{tabular}{lr}
\toprule
Backbone & \shortstack{MotionPort gain below\\25\,s/episode (pp)} \\
\midrule
HWM & +59.3 \\
H-JEPA & +54.7 \\
LeWM & +74.0 \\
\bottomrule
\end{tabular}
\caption{\textbf{Shared time allowance.} Improvements reported in
main-paper Section~4.5. Each method uses its best evaluated configuration
within the allowance; this does not require equal candidate counts.}
\label{tab:supp_shared_time}
\end{table}

MotionPort with HWM reaches 72.7\% success in approximately 4 seconds
per episode, while the best evaluated native HWM point reaches 49.3\%
at approximately 40 seconds. Section~D.4 specifies the common sequence-optimization
and execution conventions.

\subsection{Goal retention under continued feedback}
\label{sec:supp_retention_complete}\label{sec:supp_exp_retention}

\paragraph{Goal attainment and endpoint success.}
\Cref{tab:supp_retention_methods_manipulation,tab:supp_retention_methods_fourroom,tab:supp_retention_methods_ant}
extend main Figure~7 to all seven evaluation settings and all baseline
methods, including LeFlow.

\begin{table}[!tbp]
\centering\small
\begingroup
\setlength{\tabcolsep}{3pt}
\newcommand{\retentionstat}[2]{#1\,{\normalfont\scriptsize$(\pm #2)$}}
\begin{tabular*}{\columnwidth}{@{\extracolsep{\fill}}lrrr@{}}
\toprule
Method & Ever (\%) & Terminal (\%) & $\Delta$ (pp) \\
\midrule
\multicolumn{4}{@{}l}{\textit{PushT}} \\
H-JEPA & \retentionstat{46.67}{4.10} & \retentionstat{31.33}{3.81} & 15.33 \\
HWM-Macro & \retentionstat{47.33}{4.07} & \retentionstat{32.00}{3.83} & 15.33 \\
LeWM & \retentionstat{40.00}{4.03} & \retentionstat{26.00}{3.60} & 14.00 \\
SAGE & \retentionstat{44.00}{4.07} & \retentionstat{39.33}{4.01} & 4.67 \\
LeFlow & \retentionstat{17.33}{3.11} & \retentionstat{0.00}{0.00} & 17.33 \\
Hi-LeWM-C & \retentionstat{18.00}{3.15} & \retentionstat{0.67}{0.67} & 17.33 \\
MotionPort & \retentionstat{92.67}{2.13} & \retentionstat{84.67}{2.96} & 8.00 \\
\midrule
\multicolumn{4}{@{}l}{\textit{Cube}} \\
H-JEPA & \retentionstat{49.33}{4.08} & \retentionstat{36.00}{3.88} & 13.33 \\
HWM-Macro & \retentionstat{61.33}{4.00} & \retentionstat{17.33}{3.07} & 44.00 \\
LeWM & \retentionstat{30.67}{3.79} & \retentionstat{24.67}{3.54} & 6.00 \\
SAGE & \retentionstat{68.67}{3.79} & \retentionstat{66.67}{3.83} & 2.00 \\
LeFlow & \retentionstat{92.00}{0.67} & \retentionstat{0.67}{0.67} & 91.33 \\
Hi-LeWM-C & \retentionstat{4.00}{1.61} & \retentionstat{0.00}{0.00} & 4.00 \\
MotionPort & \retentionstat{97.33}{1.31} & \retentionstat{92.00}{2.20} & 5.33 \\
\bottomrule
\end{tabular*}
\endgroup
\caption{\textbf{Goal attainment and final-step success: manipulation.} Ever success and its SE reproduce main-paper Table~1. Terminal success tests the final execution step. Parentheses give the reported evaluation SEs; $\Delta=\mathrm{Ever}-\mathrm{Terminal}$ in percentage points, with rounding applied after subtraction where available.}
\label{tab:supp_retention_methods_manipulation}
\end{table}

\begin{table}[!tbp]
\centering\small
\begingroup
\setlength{\tabcolsep}{3pt}
\newcommand{\retentionstat}[2]{#1\,{\normalfont\scriptsize$(\pm #2)$}}
\begin{tabular*}{\columnwidth}{@{\extracolsep{\fill}}lrrr@{}}
\toprule
Method & Ever (\%) & Terminal (\%) & $\Delta$ (pp) \\
\midrule
\multicolumn{4}{@{}l}{\textit{FourRoom standard}} \\
H-JEPA & \retentionstat{98.00}{2.00} & \retentionstat{98.00}{2.00} & 0.00 \\
HWM-Macro & \retentionstat{98.00}{2.00} & \retentionstat{98.00}{2.00} & 0.00 \\
LeWM & \retentionstat{58.00}{7.05} & \retentionstat{58.00}{7.05} & 0.00 \\
SAGE & \retentionstat{98.00}{2.00} & \retentionstat{98.00}{2.00} & 0.00 \\
LeFlow & \retentionstat{0.00}{0.00} & \retentionstat{0.00}{0.00} & 0.00 \\
Hi-LeWM-C & \retentionstat{38.00}{6.93} & \retentionstat{10.00}{4.29} & 28.00 \\
MotionPort & \retentionstat{100.00}{0.00} & \retentionstat{100.00}{0.00} & 0.00 \\
\midrule
\multicolumn{4}{@{}l}{\textit{FourRoom hard}} \\
H-JEPA & \retentionstat{41.00}{4.94} & \retentionstat{41.00}{4.94} & 0.00 \\
HWM-Macro & \retentionstat{23.00}{4.23} & \retentionstat{23.00}{4.23} & 0.00 \\
LeWM & \retentionstat{0.00}{0.00} & \retentionstat{0.00}{0.00} & 0.00 \\
SAGE & \retentionstat{80.00}{4.02} & \retentionstat{80.00}{4.02} & 0.00 \\
LeFlow & \retentionstat{0.00}{0.00} & \retentionstat{0.00}{0.00} & 0.00 \\
Hi-LeWM-C & \retentionstat{0.00}{0.00} & \retentionstat{0.00}{0.00} & 0.00 \\
MotionPort & \retentionstat{100.00}{0.00} & \retentionstat{100.00}{0.00} & 0.00 \\
\bottomrule
\end{tabular*}
\endgroup
\caption{\textbf{Goal attainment and final-step success: fourroom.} Ever success and its SE reproduce main-paper Table~1. Terminal success tests the final execution step. Parentheses give the reported evaluation SEs; $\Delta=\mathrm{Ever}-\mathrm{Terminal}$ in percentage points, with rounding applied after subtraction where available.}
\label{tab:supp_retention_methods_fourroom}
\end{table}

\begin{table}[!tbp]
\centering\small
\begingroup
\setlength{\tabcolsep}{3pt}
\newcommand{\retentionstat}[2]{#1\,{\normalfont\scriptsize$(\pm #2)$}}
\begin{tabular*}{\columnwidth}{@{\extracolsep{\fill}}lrrr@{}}
\toprule
Method & Ever (\%) & Terminal (\%) & $\Delta$ (pp) \\
\midrule
\multicolumn{4}{@{}l}{\textit{AntMaze standard}} \\
H-JEPA & \retentionstat{78.00}{5.92} & \retentionstat{76.00}{6.10} & 2.00 \\
HWM-Macro & \retentionstat{60.00}{7.00} & \retentionstat{48.00}{7.14} & 12.00 \\
LeWM & \retentionstat{24.00}{6.10} & \retentionstat{22.00}{5.92} & 2.00 \\
SAGE & \retentionstat{20.00}{5.71} & \retentionstat{20.00}{5.71} & 0.00 \\
LeFlow & \retentionstat{2.00}{2.00} & \retentionstat{2.00}{2.00} & 0.00 \\
Hi-LeWM-C & \retentionstat{18.00}{5.49} & \retentionstat{16.00}{5.24} & 2.00 \\
MotionPort & \retentionstat{92.00}{3.88} & \retentionstat{76.00}{6.10} & 16.00 \\
\midrule
\multicolumn{4}{@{}l}{\textit{AntMaze hard}} \\
H-JEPA & \retentionstat{22.82}{4.94} & \retentionstat{22.82}{4.94} & 0.00 \\
HWM-Macro & \retentionstat{14.00}{4.29} & \retentionstat{14.00}{4.29} & 0.00 \\
LeWM & \retentionstat{3.00}{1.71} & \retentionstat{3.00}{1.71} & 0.00 \\
SAGE & \retentionstat{11.00}{3.14} & \retentionstat{11.00}{3.14} & 0.00 \\
LeFlow & \retentionstat{0.00}{0.00} & \retentionstat{0.00}{0.00} & 0.00 \\
Hi-LeWM-C & \retentionstat{0.00}{0.00} & \retentionstat{0.00}{0.00} & 0.00 \\
MotionPort & \retentionstat{58.00}{4.96} & \retentionstat{49.00}{5.02} & 9.00 \\
\midrule
\multicolumn{4}{@{}l}{\textit{AntMaze OGBench}} \\
H-JEPA & \retentionstat{0.00}{0.00} & \retentionstat{0.00}{0.00} & 0.00 \\
HWM-Macro & \retentionstat{0.00}{0.00} & \retentionstat{0.00}{0.00} & 0.00 \\
LeWM & \retentionstat{0.00}{0.00} & \retentionstat{0.00}{0.00} & 0.00 \\
SAGE & \retentionstat{4.00}{4.00} & \retentionstat{4.00}{4.00} & 0.00 \\
LeFlow & \retentionstat{0.00}{0.00} & \retentionstat{0.00}{0.00} & 0.00 \\
Hi-LeWM-C & \retentionstat{0.00}{0.00} & \retentionstat{0.00}{0.00} & 0.00 \\
MotionPort & \retentionstat{72.00}{10.20} & \retentionstat{56.00}{9.80} & 16.00 \\
\bottomrule
\end{tabular*}
\endgroup
\caption{\textbf{Goal attainment and final-step success: ant.} Ever success and its SE reproduce main-paper Table~1. Terminal success tests the final execution step. Parentheses give the reported evaluation SEs; $\Delta=\mathrm{Ever}-\mathrm{Terminal}$ in percentage points, with rounding applied after subtraction where available.}
\label{tab:supp_retention_methods_ant}
\end{table}

\paragraph{Prediction at already-successful states.}
Main Figure~7 measures the final consequence of continued feedback. To
inspect the associated prediction failure, we separately examine all
Cube decisions that start inside the 4\,cm goal region. A loss is a
prefix whose actual endpoint leaves that region. Each system contributes
600 planning decisions across 150 episodes.

The MLP and linear diagnostic position readouts agree that most native
losses were predicted to remain successful: 44/45 and 42/45. The
corresponding counts are 18/76 and 11/76 for MP-noGeo, and 0/3 for
MotionPort. Both readouts are used only for evaluation.

Across all already-successful starts, mean 3-D forecast error is
6.383\,cm for native HWM and 1.553\,cm for MotionPort. The observation
supports the connection between planning-time prediction reliability and
retention.

\paragraph{Failure cases after reaching the goal.}
\Cref{tab:supp_native_retention_cases} examines four large native
departures from the goal region. At these decisions, the hierarchy
already directs the low-level planner toward the final goal. Three of
four cases reduce predicted goal cost, although all four executions
increase realized cost and move the cube outside the success region.
The mismatch therefore persists even when the planner explicitly
optimizes for the final goal.
\begin{table*}[!tbp]
\centering\small
\setlength{\tabcolsep}{4pt}
\begin{tabular*}{\textwidth}{@{\extracolsep{\fill}}lrrrrr@{}}
\toprule
Controller & Starts inside & Leaves goal & Loss rate (\%) & \multicolumn{2}{c}{Predicts inside on a loss} \\
& & & & MLP readout & Linear readout \\
\midrule
HWM-Macro & 106 & 45 & 42.5 & 44 & 42 \\
MP-noGeo & 113 & 76 & 67.3 & 18 & 11 \\
MotionPort & 41 & 3 & 7.3 & 0 & 0 \\
\bottomrule
\end{tabular*}
\caption{\textbf{Goal departures after success on Cube with frozen HWM.} Each system has 600 planning decisions in 150 episodes. A loss starts with the real cube within 4\,cm of the goal and ends with it outside after the executed prefix. The last two columns classify predicted positions with the shared MLP and linear evaluation readouts.}
\label{tab:supp_retention_decisions}
\end{table*}

\begin{table*}[!tbp]
\centering\small
\setlength{\tabcolsep}{4pt}
\begin{tabular*}{\textwidth}{@{\extracolsep{\fill}}lrrrrr@{}}
\toprule
Example & \multicolumn{2}{c}{Goal distance (cm)} & \multicolumn{3}{c}{Final-goal cost} \\
& Before & After & Current & Predicted & Realized \\
\midrule
1 & 3.845 & 18.207 & 2.374 & 1.555 & 406.658 \\
2 & 1.098 & 14.461 & 129.582 & 15.945 & 377.804 \\
3 & 2.487 & 15.197 & 0.286 & 3.332 & 319.310 \\
4 & 1.491 & 13.975 & 11.990 & 3.277 & 227.975 \\
\bottomrule
\end{tabular*}
\caption{\textbf{Native predictions during four goal departures.}
The examples are the four largest departures among 43 eligible episodes.
All start within the 4\,cm goal region. Three have lower predicted cost
than the current state, but all four have higher realized cost after
execution. These selected failure cases illustrate the discrepancy
between imagined progress and physical outcome.}
\label{tab:supp_native_retention_cases}
\end{table*}

\section{Training and Planning Details}
\label{sec:supp_method_details}
\setcounter{equation}{0}
\renewcommand{\theequation}{D.\arabic{equation}}

This section describes how we train the motion-conditioned dynamics,
proposal priors, and temporal embedding, and how we optimize and execute
motion sequences.

\subsection{Frozen features, skill dynamics, and body propagation}
\label{sec:supp_dynamics_recipe}

The observation encoder is supplied by a released LeWM, HWM, or H-JEPA
model and remains frozen. Hierarchical backbones use features from
the active planning level with their original observation and auxiliary
inputs. Each backbone uses its own feature coordinates and normalization. The comparison across frozen backbones is in main-paper
Section~4.5.

\paragraph{Action generation from the same motion code.}
Each training window pairs the representative body trajectory with its
corresponding primitive action chunk and current body state. After fixing the
motion encoder, its standardized mean code conditions both the decoder
and the predictor. For the decoder, the flow-matching velocity target is
the difference between the normalized action chunk and a Gaussian noise
sample, with a uniformly sampled interpolation time. Integrating the
conditional velocity field yields the normalized action chunk. At
inference, fixed-step integration starts from zero noise, followed by
inverse action normalization and clipping to the environment's command
bounds. Decoding is therefore deterministic for a fixed current body
state and motion code. These numerical settings remain fixed while
planning optimizes the code. The same code conditions $G$ directly;
it is not obtained by re-encoding the decoded action trajectory.

\paragraph{Prefix prediction.}
The skill predictor uses a four-hidden-layer residual MLP of width 1,024
with SiLU activations. One shared trunk predicts visual-feature residuals
at all prefixes. Inputs and residual targets are standardized with fixed
training-window means and population standard deviations. The predicted
residuals are transformed back and added to the current feature, so the
output remains in the released encoder's coordinates. The prediction
loss is mean squared error over samples, prefixes, and feature dimensions.

The prefix set is $\{5,10\}$ for PushT and $\{5,10,20\}$ for the other
three domains. Cube and Ant use a separate residual body-state head,
trained against the future body state with a standardized error
weighted by 0.1. PushT propagates pusher position and velocity through its
known control law applied to decoded, bounded commands. FourRoom uses no
separate body-state input. These alternatives supply the predicted body state used by the next motion in the main rollout.

The main configuration learns single-skill prefix effects and recursively
composes them at planning time, using predicted visual and body states.
Training uses AdamW with learning rate $3\times10^{-4}$,
weight decay 0.01, batch size 1,024, 500 warm-up updates, and cosine decay
to 3\% of the peak learning rate. We train for 100,000
updates on PushT and Ant, 60,000 on Cube, and 50,000 on FourRoom,
using the final model in each case.

\subsection{Conditional flow priors}
\label{sec:supp_prior_recipe}

Each window contributes the standardized mean code of its frozen motion
encoder. The goal-conditioned prior receives the current visual feature,
the current body state where used, and a future feature from the same
episode. Its goal offset is sampled uniformly from $L$, $3L$, $5L$, and
the episode end; the first three offsets are clipped to the last available
frame. The goal-free prior omits the goal feature. Its supervised target
is the same one-skill motion code. $G$ connects these one-skill proposals into sequences at planning time.

Each prior is a conditional flow network with three width-256 SiLU hidden
layers and a 16-dimensional sinusoidal flow-time embedding. Training uses
independent Gaussian noise and uniform flow time, and regresses the
noise-to-code displacement. Codes and conditioning inputs have fixed
training-set normalization. Sampling uses eight Euler steps from Gaussian
noise; the output normalization is inverted to recover the motion
encoder's standardized planning coordinates.

Both priors use AdamW with learning rate $10^{-3}$, weight decay 0.01,
batch size 1,024, and 30,000 updates. We use a fixed episode-level
development split and evaluate its loss every 2,000 updates. The
trained prior with the lowest development loss is used for evaluation.
Goal-conditioned and goal-free priors are trained independently; their
relative allocation within the learned component is fixed for the evaluation configuration. In the main prior comparison, the learned component and Gaussian proposals each receive half of the initial candidates. Goal-conditioned and goal-free draws use the same one-motion target and the same planning coordinates.

\subsection{Temporal embedding and goal relabeling}
\label{sec:supp_temporal_recipe}

The temporal embedding has two independently parameterized MLP heads.
Each standardizes its input with training-node statistics, applies two
width-256 linear layers with LayerNorm and SiLU, and outputs a
32-dimensional vector through a bias-free linear layer. Its value is
the negative distance between current-state and goal embeddings.
For numerical stability, the implementation evaluates
$-\sqrt{\max(\|\phi(h)-\phi(h_g)\|_2^2,10^{-6})}$.
The planner averages the two positive distances.

Temporal learning uses transitions separated by 20 environment
steps. Goals are sampled as the transition's current node, a uniformly
sampled strictly later node of the same trajectory, or a random node
of the same training split, with probabilities 0.2, 0.5, and 0.3.
The reward is zero when the current and goal nodes are identical and
$-1$ otherwise, with discount 0.99 per training edge.
Transition and goal samples come from the offline training split.

Each embedding head uses its corresponding target head's
bootstrapped value as that head's regression target. The minimum of the
two targets is used only to compute the shared advantage against the
mean current target value. Both heads then receive the same expectile
weight: 0.95 for nonnegative advantage and 0.05 otherwise. The two
weighted squared errors are summed over heads and averaged over the
minibatch. Targets and advantage weights are computed without gradients.

Only the online embedding parameters are optimized. Target parameters
are updated after every optimizer step as
\begin{equation}
 \bar\nu\leftarrow(1-\eta_{\rm EMA})\bar\nu+\eta_{\rm EMA}\nu,
 \qquad \eta_{\rm EMA}=0.005.
 \label{eq:supp_phi_ema}
\end{equation}
Training uses AdamW with learning rate $3\times10^{-4}$, weight decay
$10^{-4}$, gradient-norm clipping at 1, and batch size 128.
We use the final checkpoint after 6,000 training updates. Each head
uses the feature coordinates and normalizer of its backbone.

The learned distance measures discounted temporal reachability at the
training transition stride. Current-goal relabeling anchors the value,
future and random goals supply reachability relationships, and TD
bootstrapping propagates them through the observed transition graph.

\subsection{Sequence optimization and execution conventions}
\label{sec:supp_optimizer}

For each planning call, the observed starting state is encoded once and
held fixed during optimization. Alternating one-skill prior sampling and
skill-dynamics rollout produces the initial candidates
$Z_i^{(0)}=(z_{i,1}^{(0)},\ldots,z_{i,H}^{(0)})$. The starting observed features and all learned module weights remain fixed during this optimization.

\begin{samepage}
AdamW jointly updates each candidate sequence:
\begin{equation}
 Z_i^{(r+1)}=\operatorname{AdamWStep}\!\left(
 Z_i^{(r)},\nabla_Z J_i(Z_i^{(r)})\right).
 \label{eq:plan_update}
\end{equation}
\end{samepage}
The optimizer's momentum and second-moment state are initialized anew
for each planning call. Gradients pass through the composed skill
predictor and body-state updates. The prior is not resampled during
these updates. AdamW refines unconstrained motion codes, while decoded
commands are clipped to the environment bounds.

\Cref{tab:supp_optimizer_settings} summarizes the shared optimizer
and stopping settings. The planning-budget study in Section~C.6 varies
candidate and update budgets while holding the trained models fixed.
PushT, Cube, and FourRoom use the final optimization iterates; AntMaze
uses each candidate's lowest-cost iterate, including its initialization.

\begin{table}[!tbp]
\centering\small
\setlength{\tabcolsep}{3pt}
\begin{tabular*}{\columnwidth}{@{\extracolsep{\fill}}lr@{}}
\toprule
Setting & Value \\
\midrule
Optimizer & AdamW \\
Weight decay & 0.01 \\
Minimum / maximum updates & 30 / 90 \\
Early-stopping patience & 5 updates \\
Relative improvement threshold & 1\% \\
\bottomrule
\end{tabular*}
\caption{\textbf{Shared motion-sequence optimization settings.}
An improvement exceeding 1\% in the minimum candidate objective resets
the patience counter. Early stopping begins after 30 updates.}
\label{tab:supp_optimizer_settings}
\end{table}

Visual and temporal goal-distance normalization constants remain fixed
throughout optimization.

The objective combines goal costs along the predicted sequence with
normalized predictor disagreement, as defined in main-paper Section~3.3.
The two dynamics models evaluate the same motion sequence; their feature
discrepancy is divided by a fixed calibration scale. When the remaining
environment budget ends within a motion segment, the objective uses the
latest predicted prefix within that budget.

The lowest-cost plan supplies the common first-skill subgoal.
First skills from the top candidates are decoded from the same
current body state and checked by the frozen native predictor, with
feature and temporal projection into the planning coordinates. This
check evaluates the mean squared feature distance to the common
first-motion subgoal without gradients. It reselects a decoded candidate and does not run
another low-level optimizer. The agent executes the selected command
prefix and replans from a fresh observation.

\section{Limitations}
\label{sec:supp_limitations}

\paragraph{Dependence on the motion interface.}
MotionPort uses body-motion similarity to organize planning, so its
benefit depends on how much physical structure the original action
representation already provides. In PushT, native commands already
closely describe pusher movement, and geometry regularization offers
limited improvement in pairwise motion-distance alignment. Gains from
individual components also vary across settings; their strongest effect
comes from their combination in the planning system.

\paragraph{Offline coverage and temporal scoring.}
The learned motions, priors, and temporal embedding depend on the
movements and transitions represented in the offline data. The temporal
score improves global goal-distance ranking, but can remain locally
misleading.
Extending coverage and learning more accurate local reachability models
are promising directions for more complex manipulation and navigation.

\end{document}